\documentclass[twoside,journal,comsoc]{IEEEtran}
 
\usepackage[pdftex]{graphicx}
\usepackage{floatrow}
\usepackage[caption=false,font=footnotesize]{subfig}

\usepackage{tikz} 
\usepackage{scalerel}
\usetikzlibrary{svg.path}
\definecolor{orcidlogocol}{HTML}{A6CE39}
\tikzset{
    orcidlogo/.pic={
        \fill[orcidlogocol] svg{M256,128c0,70.7-57.3,128-128,128C57.3,256,0,198.7,0,128C0,57.3,57.3,0,128,0C198.7,0,256,57.3,256,128z};
        \fill[white] svg{M86.3,186.2H70.9V79.1h15.4v48.4V186.2z}
        svg{M108.9,79.1h41.6c39.6,0,57,28.3,57,53.6c0,27.5-21.5,53.6-56.8,53.6h-41.8V79.1z M124.3,172.4h24.5c34.9,0,42.9-26.5,42.9-39.7c0-21.5-13.7-39.7-43.7-39.7h-23.7V172.4z}
        svg{M88.7,56.8c0,5.5-4.5,10.1-10.1,10.1c-5.6,0-10.1-4.6-10.1-10.1c0-5.6,4.5-10.1,10.1-10.1C84.2,46.7,88.7,51.3,88.7,56.8z};
    }
}
\newcommand\orcidicon[1]{\href{https://orcid.org/#1}{\mbox{\scalerel*{
                \begin{tikzpicture}[yscale=-1,transform shape]
                \pic{orcidlogo};
                \end{tikzpicture}
            }{|}}}}

\usepackage[english]{babel}
\usepackage[T1]{fontenc}
\usepackage{cite}
\usepackage[hidelinks,
            pdfencoding=auto,
            pdfauthor={Luca Buoncompagni, Syed Yusha Kareem, and Fulvio Mastrogiovanni},
            pdftitle={Human Activity Recognition Models in Ontology Networks},
            pdfsubject={Activity model semantically contextualised and scheduled by a smart home (DOI: 10.1109/TCYB.2021.3073539).},
            pdfkeywords={Activities of Daily Living, Smart Environment, Spatial-Temporal Reasoning, Ontology Design, Knowledge Representation Formalisms and Methods, Semantic networks, Pattern Recognition, Explainable AI.}]{hyperref}
\usepackage{xspace}
\usepackage{multirow}
\usepackage{amsmath}
\input{ushyphex}
\usepackage{mathtools}
\usepackage{bm}
\usepackage[cmintegrals]{newtxmath}
\usepackage{mathrsfs}

\newcommand{\MON}{Arianna\texorpdfstring{\textsuperscript{+}}{\textplussuperior}\xspace}
\newcommand{\eg}{\emph{e}.\emph{g}.,~} 
\newcommand{\ie}{\emph{i}.\emph{e}.,~}
\newcommand{\dom}[1]{\ensuremath{\mathbb{#1}}\xspace}
\newcommand{\staS}[1]{\ensuremath{s_{#1}}\xspace}
\newcommand{\staT}[1]{\ensuremath{t_{#1}}\xspace}
\newcommand{\dltext}[1]{{\fontfamily{lmss}\selectfont {#1}}}
\newcommand\mcirc[1]{\ensuremath{\circ_{_{\!#1}}}\xspace}
\newcommand{\PAR}[1]{\subsubsection{#1}}
\newcommand{\tdl}[1]{\text{\dltext{#1}}\xspace}
\newcommand{\expl}[1]{\begin{center}\emph{``#1''}\end{center}}
\newcommand{\vs}[1]{\ensuremath{\widetilde#1}}
\newcommand{\dlsub}[2]{\ensuremath{\tdl{#1}_\tdl{#2}}}

\begin{document}
\title{Human Activity Recognition Models\\in Ontology Networks${^\star}$}%
\author{Luca~Buoncompagni\textsuperscript{\orcidicon{0000-0001-8121-1616}},~%
        Syed~Yusha~Kareem\textsuperscript{\orcidicon{0000-0002-2360-1680}},~%
        and~Fulvio~Mastrogiovanni\textsuperscript{\orcidicon{0000-0001-5913-1898}}%
\thanks{All the authors are affiliated with the Department of Informatics, Bioengineering, Robotics and Systems Engineering, University of Genoa, Via Opera Pia 13, 16145, Genoa, Italy.}%
\thanks{Luca~Buoncompagni and Fulvio~Mastrogiovanni are also affiliated with Teseo s.r.l., Piazza Montano 2a,  16151 Genoa, Italy.}%
\thanks{Corresponding author's email: \href{mailto:luca.buoncompagni@edu.unige.it}{luca.buoncompagni@edu.unige.it}.}%
\thanks{\par\vspace*{-.7em}$^\star$%
Manuscript submitted to the \emph{IEEE Transactions on Cybernetics} journal on February 1, 2020; revised on October 18 and on February 2; accepted on March 30, 2021, with Digital Object Identifier (DOI) \href{http://dx.doi.org/10.1109/TCYB.2021.3073539}{10.1109/TCYB.2021.3073539}. This document contains the accepted version of the manuscript.}
}

\newcommand{\KL}[1]{{#1}}
\markboth{H\KL{uman} A\KL{ctivity} R\KL{ecognition} M\KL{odels in} O\KL{ntology} \KL{Networks}}%
         {L\KL{uca} B\KL{uoncompagni}, S\KL{yed} Y\KL{usha} K\KL{areem}, \KL{and} F\KL{ulvio} M\KL{astrogiovanni}}

\maketitle
\thispagestyle{empty}

\begin{abstract}
We present \MON, a framework to design networks of ontologies for representing knowledge enabling smart homes to perform human activity recognition online.
In the network, nodes are ontologies allowing for various data contextualisation, while edges are general-purpose computational procedures elaborating data.
\MON provides a flexible interface between the inputs and outputs of procedures and \emph{statements}, which are atomic representations of ontological knowledge.
\MON schedules procedures on the basis of \emph{events} by employing logic-based reasoning, \ie by checking the classification of certain statements in the ontologies.
Each procedure involves input and output statements that are differently contextualised in the ontologies based on specific prior knowledge.
\MON allows to design networks that encode data within multiple contexts and, as a reference scenario, we present a modular network based on a spatial context shared among all activities and a temporal context specialised for each activity to be recognised.
In the paper, we argue that a network of small ontologies is more intelligible and has a reduced computational load than a single ontology encoding the same knowledge.
\MON integrates in the same architecture heterogeneous data processing techniques, which may be better suited to different contexts.
Thus, we do not propose a new algorithmic approach to activity recognition, instead, we focus on the architectural aspects for accommodating logic-based and data-driven activity models in a context-oriented way.
Also, we discuss how to leverage data contextualisation and reasoning for activity recognition, and to support an iterative development process driven by domain experts.
\end{abstract}


\section{Introduction}
\label{sec:introduction}

\IEEEPARstart{T}{he} paper concerns an architecture for human activity recognition that is able to detect and classify multiple activity patterns from streaming sensory data under the following requirements.
\emph{(i)} The architecture should be \emph{intelligible} since the explanation of why an activity pattern is (or is not) recognised is one of its outcomes.
This is crucial if domain experts can only provide qualitative or partial knowledge of the patterns to be recognised.
As a consequence, \emph{(ii)} architecture designers and domain experts must establish a close collaboration to attain an effective detection and classification mechanism.
In order to facilitate this collaboration, we adhere to an iterative development process~\cite{larman2004agile} involving different runs of the develop-test-improve phases.
Hence, \emph{(iii)} such features as \emph{modularity} and \emph{flexibility} of the architecture are crucial for limiting the costs in terms of design effort, data collection, and computational load.
Finally, \emph{(iv)} the recognition of activity patterns may require a certain reaction by the architecture, and therefore recognition should be done \emph{online}.

We consider a scenario whereby sensors distributed in an environment feed data into algorithms to assess the well-being of elderly people who live independently and are, to a certain extent, autonomous.
Conventionally, such assessment is usually performed by caregivers and geriatricians via a qualitative evaluation of the Activities of Daily Living (ADL)~\cite{Katz59adl}, which is used by such domain experts to analyse an elderly's lifestyle evolution by looking at their routine.
ADL are usually decomposed into sub-activities, which can be analysed in terms of gestural movements, poses and postural transitions, as well as using a series of cognitive dimensions such as activity planning.
Since this evaluation is qualitative, it can vary based on the person being monitored and the caregiver's experience. 
As a motivation for the work discussed in this paper, we foresee activity recognition approaches developed in close collaboration with caregivers and geriatricians to support a qualitative evaluation of ADL with explainable outcomes in terms of \emph{behavioural analysis}.     

\IEEEpubidadjcol

From a medical perspective, there is a lack of evidence about the technology readiness level of smart home architectures supporting elderly and people with special needs by recognising ADL~\cite{liu2016medicalsurvay}.
We hypothesise that this may be due to a lack of effective collaboration between technical and domain experts, especially when considering the recognition and classification of human activities. 
In this paper, we present a framework, which we refer to as \MON, designed to meet the requirements highlighted above.
\MON allows to design ontology networks, whereby ADL are decomposed into sub-activities processed by a number of modular, high-level symbolic reasoners, which assures intelligibility (in support of domain experts), as well as flexibility (to support iterative development processes).
Differently from other \emph{architectural} approaches, we use multiple contexts to encode the knowledge required for recognising and classifying ADL instead of using a single representation.
\MON includes a general-purpose interface between data, which is contextualised based on prior knowledge, and algorithms.
This enforces an approach that limits the data representation complexity, and allows to accommodate different classes of algorithms for data processing.

The paper discusses a number of features related to intelligibility and flexibility exhibited by \MON on the basis of a possible ontology network for smart homes, and the discussion is supported using a popular ADL-related dataset. 
We present a modular ontology network design that involves a spatial representation concerning all the activities to be recognised, and a temporal representation specified for each activity.
With this approach, the computation load of \MON with respect to other approaches using a single ontology for data contextualisation is reduced because we reason on smaller structures, on average, and only when required.
Also, we argue that many, small ontologies are more intelligible since they are simpler in terms of classes and relationships.
Furthermore, we observe that a hierarchy of ontologies does not jeopardise the activity recognition and classification rates compared to systems using a single data representation structure.

It is worth mentioning that we neither want to investigate specific types of activities, nor compare our approach with other \emph{algorithms} for activity recognition and classification models.
On the contrary, we posit that \MON allows to design architectures that accommodate heterogeneous, event-based, data processing techniques for human activity recognition and classification.

\section{Related Work}
\label{sec:relatedwork}

From an analysis of the literature, it emerges that there are many techniques for recognising the same human activities by employing different sensors and, as discussed in~\cite{amiribesheli2015survay2}, each of them has benefits and limitations. 
The type of sensors used in human activity recognition and classification in smart homes generate spatial-relevant information about occurring \emph{events} (\eg passive infrared sensors distributed in the environment for motion detection), or provide a continuous \emph{stream} of raw data, (\eg wearable inertial sensors or cameras).

Models to recognise human activities have been formalised using two classes of techniques.
($i$)~Data-driven approaches, in turn, divided into the Machine Learning (ML) (\eg Artificial Neural Networks~\cite{Begg2006} or Support Vector Machines~\cite{Fleury2010}), and the probabilistic (\eg Hidden Markov Models~\cite{casasDataset} and the Bayesian Networks~\cite{nazerfard2013}) based approaches. 
($ii$)~Logic-based approaches to reason on contextualised data based on knowledge available \emph{a priori}.
In the last case, the Ontology Web Language (OWL) standard has been widely proposed for human activity recognition, as discussed in~\cite{abburu2012reasoners}.
Typically, both probabilistic and logic-based approaches exploit Allen's Interval Algebra~\cite{allen1983} for temporal representations of knowledge.

Description Logic (DL)~\cite{baader2003description} is a state of the art formalism that allows to represent and reason on symbolically structured knowledge.
DL lies within the decidable fragment of languages that are based on first-order logic, and it has been standardised by the World Wide Web Consortium (W3C) as OWL-DL.
OWL~\cite{OWL} allows to design ontologies, which are semantic \emph{corpora} containing knowledge encoded as logically-defined symbols meant at being readable by humans and machines.
An ontology is defined within a \emph{context}, \ie a consistent corpus of knowledge representing a domain of interest~\cite{Dey2001}. 
Within a context, a \emph{reasoner} can evaluate the consistency of knowledge by performing subsumption, instance-checking and inference based on well-defined rules~\cite{meditskos2016survay}. 
Several reasoners have been proposed, some of which are discussed in~\cite{abburu2012reasoners}.
Deterministic DL reasoners (\eg Pellet~\cite{pellet}) do not represent uncertainties, whereas fuzzy~\cite{f2016fuzzya} or probabilistic~\cite{r2016Probabilistica} OWL reasoners do.
DL reasoning is computationally expensive, and this limits its online usage, especially when the context is information-rich.
In particular, the computation scales exponentially with the amount of knowledge encoded in an ontology~\cite{baader2003description}. 

Data-driven techniques are robust to uncertainties and are amenable for online use.
They can be employed with sensory data streams \cite{carfi2018online}, and with event-based data \cite{Fleury2010}.
Although data-driven techniques do not need extensive knowledge engineering as it is required by logic-based approaches, they involve much effort in data collection, which may be still problematic for effective human activity recognition in smart homes.
Furthermore, they exhibit weak properties in explainability and intelligibility but they might notably improve the activity recognition accuracy of an architecture.
ML could effectively be used to generate \emph{discrete} events by processing a sensory data stream, \eg the occurrence of gestures using wrist-related inertial data \cite{carfi2018online}, whereas probabilistic methods could estimate sub-activity models of ADL \cite{civitarese2019newnectar}.
\MON integrates DL to assure intelligibility for domain experts, and this paper does not rely on data-driven techniques but rather uses simple logic-based activity models.
This is because we focus on highlighting the architectural aspects of \MON rather than on the recognition rates of an ontology network developed based on \MON.
Nonetheless, the paper discusses how \MON can accommodate general-purpose computational procedures, including data-driven models, based on complementary contexts.

Other hybrid logic-statistical frameworks have been proposed in the literature. 
For instance, in~\cite{helaoui2013multi,riboni2016multi,matassa2019reasoning} activities are described with a combination of probabilistic models and semantic constraints.
Other contextualisation techniques exploit a hybrid solution via the augmentation of ontologies with specific ML-based algorithms, which reason in terms of activity patterns~\cite{chen2014hybrid,mojarad2018hybrid}, or infer concepts through a learning approach~\cite{salguero2019ontology}.
In~\cite{salguero2018ontology}, an ontology is used to identify features useful for training human activity models, whereas for the same purpose a probabilistic ontology is used in~\cite{helaoui2013multi}.
Markov Logic Networks (MLN) are used in~\cite{bodart2014owlMLN}, are extended by Allen's Algebra in~\cite{gayathri2017ontoML}, and are adopted in~\cite{civitarese2019newnectar} to refine structures designed through knowledge engineering practice. 
Differently from these approaches, \MON allows an arbitrary procedure to concurrently interface with different knowledge \emph{corpora} whose structure is \emph{a priori} defined.

Context-aware approaches have been shown to be effective in evaluating human activities if such evaluation is performed on data semantically structured in an ontology.
The effectiveness depends on the prior knowledge used to structure data, and this leads to a knowledge engineering issue.
Indeed, finding the conceptualisation that best describes the context to support the recognition of human activities is far from trivial.
Several ontology-based representations have been evaluated in~\cite{salguero2019ontology} on the basis of the context they define, \eg sensor events, sensor hierarchies, human postures or locations.
In \cite{matassa2019reasoning}, the context is defined in terms of \emph{affordances}, and it is based on relevant knowledge about space, time and human attitudes.
In these approaches, the need for efficient reasoning on data typically leads to ontologies that are engineered either for computational efficiency or expressivity, \ie they precisely describe a human activity in detail.
For instance, the work in~\cite{scalmato2013owl} reports an ontology with limited expressivity, based on OWL-EL, which has been deployed in a real-world use case.
That work is an example showcasing a trade-off between the computational load while reasoning and the expressivity of knowledge representation.

In this paper, we want to extend context-aware architectures for human activity recognition and classification based on OWL with the capability of reasoning on a set of ontologies interconnected in a network.
Being interconnected, the data represented in an ontology defining a particular context can be exchanged with another ontology representing a different context, \ie having different prior knowledge, which might be more appropriate for recognising a specific activity.
We argue that having multiple ontologies, each with a dedicated reasoner, can reduce the overall reasoning load with respect to an approach encoding all the required knowledge in a single ontology, without limiting the inference capabilities.
This is especially relevant when each ontology is related to a detailed -- but very partial -- representation of the context, \ie it is small in terms of classes and relationships.
Having several ontologies allows to further investigate the trade-off between performance and expressivity since the knowledge in each ontology can be hierarchically structured and concurrently evaluated.

Other context-aware reasoning techniques have been proposed to recognise ADL based on sub-activities through networks based on the Dempster-Shafer theory of evidence~\cite{hong2009ontologyNetwrok}. 
A hierarchical activity recognition approach has been proposed in~\cite{gayathri2015hierarchicalAR} for a reactive system based on layers evaluated by an MLN to describe abnormalities.
The detection of abnormal behaviour has been proposed as an effective context to represent outcomes relevant for assistance application \cite{matassa2019reasoning}.
In this case, the architecture should maintain a representation of data that is able to adapt to the routine of the assisted person.
Therefore, it may require a sort of \emph{autobiographical} memory which, in our scenario, should be intelligible for domain experts.
If a declarative type of memory is used, its contents would be more intelligible, but it would involve prior knowledge for contextualising experiences, which opens challenging issues.

\section{Contribution}
\label{sec:contribution}

Section~\ref{sec:introduction} motivates our goal to recognise human activity with the following requirements.
We require an architecture that 
($i$)~is \emph{flexible} enough to accommodate for different techniques, \eg logic-based or data-driven models,
($ii$)~is \emph{modular} to support an iterative development process engaging technical and domain experts, 
($iii$)~has outcomes that are \emph{intelligible} to domain experts, and
($iv$)~can process data \emph{online}.

We present and discuss \MON, a framework to design and deploy hybrid context-based architectures  for recognising and classifying human activities in smart homes based on a network of ontologies.
\MON reasons on contextualised representations to detect events, which would schedule computation procedures.
Contextualised representations are based on prior knowledge and sensory data, and they are encoded in OWL ontologies.
Computational procedures can implement arbitrary algorithms for aggregating sensory data in model-based sub-activities, which can be detected and classified.

We designed a general-purpose interface to share knowledge between ontologies and computational procedures, so that \MON can be \emph{flexible} with respect to different methodologies for processing raw data and reasoning on sub-activities.
This paper shows that such an interface allows to design \emph{modular} smart home architectures. 
Furthermore, the paper shows that if each ontology in the network is concerned with a well-defined, small context, then the benefits are two-fold.
($i$)~The knowledge \emph{intelligibility} increases since each ontology addresses a limited and self-contained portion of knowledge that is affected by the reasoning process.
Thus, domain experts can focus on specific aspects of the application, \eg modelling a given activity independently from the others. 
($ii$)~The reasoning load for data contextualisation is limited, and the architecture can be used \emph{online}. 
This is because each ontology has a dedicated reasoner that evaluates only the data relevant for a certain context, \eg a particular activity.
In addition, if the reasoners (and the other procedures) are invoked only when required by the means of well-contextualised events, the computation would benefit and the flow of statements among ontologies would be of immediate comprehension%
\footnote{A video, available at \url{https://youtu.be/SEEqSawrQNo}, shows through simple scenarios the intelligibility of DL symbols in a network of ontologies.}.

For a scenario based on simple fluent models of ADL and a popular dataset involving event-based sensory data (presented in Section~\ref{sec:experiment}), this paper shows how \MON can be flexibly used to develop an ontology network.
The paper also reports the recognition and classification rates obtained through logically represented symbols, and compares them with other state of the art activity models validated with the same dataset.
The comparison is done between systems that use a single representation domain (symbolic, probabilistic or hybrid) to encode the whole knowledge required to recognise human activities in the dataset, and an ontology network that involves distributed symbolic domains, \ie contexts.
The comparison shows that an ontology network can have recognition and classification performance that is comparable to state of the art techniques that are most often (to the best of our knowledge) based on a \emph{one size fits all} approach to represent data. 

Collecting data related to ADL often implicitly involves scripted behaviours~\cite{krishnan2014activity}.
Also, the dataset we use collects limited observations with respect to the whole spectrum of possible ways in which the considered activities can be performed.
Thus, the paper does not characterise the recognition and classification rates of the fluent models used in the ontology network we developed for the dataset.
In a less controlled scenario, data-driven approaches might be more accurate than our fluent models.
Moreover, to characterize our hierarchy of fluent models, we should also compare it with other logic-based representations of ADL, but this is out of the scope of this paper. 
Nevertheless, fluent models allow us to present an ontology network developed with \MON, and to discuss that it can accommodate the outcomes of both logic-based and data-driven models in a context-oriented, hierarchical way.
Therefore, this paper focusses on the problem of structuring knowledge, from an architectural perspective, such that it can be used by general-purpose procedures that require and produce intelligible data in a modular and flexible manner.

\section{An Overview of \MON}
\label{sec:overview}

\MON supports the design of smart home architectures for recognising ADL under the requirements stated in Section~\ref{sec:contribution} and motivated in Section~\ref{sec:introduction}.
It is based on a general-purpose definition of symbolic \emph{statements}, which define the atomic knowledge that \MON can process.
Statements have an associated Boolean state and a timestamp.
They can be classified in ontologies over time based on \emph{a priori} knowledge as encoded in ontologies themselves.
In \MON, each relevant piece of knowledge is modelled as a statement.
For instance, a statement can represent the fact that that a \emph{cabinet's door is open}, whereas before \emph{it was closed}, which might lead to different conclusions for different activities. 

\MON allows to design and maintain a network of ontologies, where \emph{nodes} in the network are OWL ontologies, and \emph{edges} therein are computational procedures.
Each computational \emph{procedure} implements an arbitrary algorithm that combines statements to generate new statements.
To this aim, we need to formalise a (communication) interface between computational procedures and ontologies.
The more such an interface does not constrain the behaviour of algorithms encoded as computational procedures, the better it is for an iterative development process.
In particular, we assume that a computational procedure implements an algorithm that 
($i$)~is triggered based on \emph{events} occurring in an ontology, 
($ii$)~retrieves statements from given ontologies, 
($iii$)~performs a computable sequence of steps to aggregate knowledge in the form of statements, and 
($iv$)~generates statements -- possibly -- to be stored in other ontologies.
If these conditions are met, we consider that the algorithm satisfies the knowledge \emph{interface} of \MON.
It is noteworthy that \MON by design does not distinguish \emph{basic} statements generated directly using sensory data from \emph{aggregated} statements generated by (a possibly long sequence of) computational procedures.

\MON implements a scheduler to contextualise statements using the prior knowledge encoded in different ontologies of the network and to trigger corresponding computational procedures.
A detailed account of how the scheduler works is out of the scope of this paper. 
However, it would suffice to say that its behaviour is based on how events are classified as statements in a given ontology of the network.
A procedure is not only defined in terms of input and output statements, but also with the semantic associated to events characterising the particular context where it is computed, \ie an ontology.
As a consequence, statements are subject to contextualisation, which in turn can generate events triggering further computational procedures.

\MON is provided with an upper ontology defining ontologies and procedures that would be bootstrapped in the nodes and the edges of the network.
In this paper, we present and discuss a network of ontologies able to spatially relate sensory data streams with the topological locations where the activities generating those data are supposed to be performed.
Spatial information about the environment of interest is maintained within a purposely design ontology in the network.
When an assisted person is in a given area or location, an event occurs and a procedure is scheduled to 
($i$)~select relevant spatially contextualised statements from such an ontology, 
($ii$)~use these statements to aggregate new, procedure-specific statements, and 
($iii$)~store the results in an activity ontology, \ie a temporal context.
When a new statement is introduced in the temporal representation, an event might be detected and a procedure would be triggered to evaluate a \emph{fluent model} and, eventually, recognise the activity.

Using this computational workflow, technology and domain experts can prototype, integrate, and evaluate heterogeneous techniques to activity modelling and recognition in a modular, flexible, intelligible and computationally efficient way. 
The paper focuses on the management of orchestrating events, and on the general-purpose statement representation.
However, in order to use an ontology network in real-world scenarios, different approaches should be considered to implement activity models.
Furthermore, different knowledge representation to structure ontologies should be evaluated, and the contexts represented therein validated. 
However, the paper does not focus on these aspects, whereas it proposes a preliminary implementation of a network of ontologies for smart homes only for the sake of argument.
In the next Sections we highlight the \MON's features enabling the design of activity recognition models. 

\section{Knowledge Representation}
\label{sec:statement}
\subsection{Statement's Algebra}
\label{sec:algebras}
We define a set $\dom{X}$ of statements, where each \emph{statement} $X\in\dom{X}$ represents atomic knowledge about a context of interest. 
A statement $X$ is characterised by a Boolean state $\staS{X}\in\dom{B}$ observed at a time instant $\staT{X}$.
Therefore, a statement $X$ can be defined as a tuple such that ${X = \langle \staS{X}, \staT{X} \rangle}$, where $\staT{X}\in\dom{N}^+$ is a timestamp.
In the notation we adopt, the name of a statement is always expressed in both the elements of the tuple, although we will discuss state and time independently if necessary.
However, it is important to highlight that in \MON one element of the tuple cannot exist without the other, since each statement is considered as atomic knowledge in each node of the ontology network. 

We evaluate combinations of statements using a higher-order function $\bm{f}$, which is composed of a \emph{logic} function $\bm{s}$ and an \emph{algebraic} function $\bm{t}$. 
The function $\bm{f}$ maps a set of statements ${\chi=\{X_1,\ldots, X_n\}}$ to a new aggregated statement \vs{Z}. 
Please note that we always denote statements with Roman uppercase letters and functions with italic Roman lowercase letters.
The higher-order functional $\bm{f}$ can be defined as
\begin{equation}
    \label{eq:stF}
    \begin{array}{cc}
        \begin{aligned}
            \bm{f} = \langle \bm{s}, \bm{t} \rangle \colon~~ \dom{X}^n &\to \dom{X}\\
            \chi &\mapsto \vs{Z},
        \end{aligned}
        &
        \left\{
        \begin{aligned}
            \bm{s} \colon~~ \dom{X}^n &\to \dom{B}\\[-.4em]
            \chi &\mapsto \staS{\vs{Z}},\\[.4em]
            \bm{t} \colon~~ \dom{X}^n &\to \dom{N}^+\\[-.4em]
            \chi &\mapsto \staT{\vs{Z}}.
        \end{aligned}        
        \right.
    \end{array}
\end{equation}

\noindent
In~\eqref{eq:stF}, $\bm{f}$ is treated as an \emph{aggregator function}, where \vs{Z} is a new statement having state
$\staS{\vs{Z}}=\bm{s}(\chi)$, 
and time instant 
$\staT{\vs{Z}}=\bm{t}(\chi)$.
In the special case when $n=1$, we reduce the function to a single statement's definition, \ie $\bm{s}(X_1)=\staS{X_1}$, and $\bm{t}(X_1)=\staT{X_1}$, and therefore $\bm{f}(X_1)=X_1$.

In \MON, the set of statements $\chi$ is considered to be the solution of a query to an OWL reasoner associated with an ontology in the network, \ie the ontology returns a set of statements contextualised with a common semantic (\eg all the statements related to the sensors attached to the refrigerator in a day period).
The result of applying $\bm{f}$ is a statement in an ontology, which belongs to \dom{X}.
In order to clearly distinguish raw statements from aggregated ones, we denote aggregated statements with a \emph{tilde} symbol above them, as in \vs{Z}.
We consider raw statements to originate from real physical sensors, and therefore a statement that is a result of $\bm{f}$ on a set of raw statements is considered to be an \emph{aggregated} statement.

The function $\bm{f}$ aggregates statements based on the kind of operators used between them. 
As an example, assuming $n=2$, \ie $\chi=\{X, Y\}$, we present the logical \emph{and} operator in \eqref{eq:operator1}, and the \emph{precedence} operator in \eqref{eq:operator2}
\begin{align}
    \vs{Z} \vDash X \land Y
    &\iff \bm{f} : 
    \begin{cases}
        \bm{s}{\,:}& \staS{\vs{Z}} = \staS{X} \land \staS{Y},\\
        \bm{t}{:}& \staT{\vs{Z}} = \max\{\staT{X}, \staT{Y}\}.
    \end{cases}
    \label{eq:operator1}\\
    \vs{Z} \vDash X \leqslant Y
    &\iff \bm{f} : 
    \begin{cases}
        \bm{s}{\,:}& \staS{\vs{Z}} = \top \iff \staT{X} \leqslant \staT{Y},\\
        \bm{t}{:}& \staT{\vs{Z}} = \max\{\staT{X}, \staT{Y}\}.
    \end{cases}
    \label{eq:operator2}
\intertext{%
In \eqref{eq:operator1} and in the following discussion we use the symbols $\top$ to indicate \emph{true}, and $\bot$ to denote \emph{false}, as it is customary in DL-based formalisms.
Given a set $\chi=\{X\}$, a Boolean state $\phi\in\dom{B}$, and a time instant $\delta\in\dom{N}^+$, we observe in \eqref{eq:operator3} a specification of the logical \emph{and} ($\wedge$), and a definition of the mathematical operator $+$ for time computation purposes in \eqref{eq:operator4}:}
    \vs{Z} \vDash X \land \phi
    &\iff \bm{f} : 
    \begin{cases}
        \bm{s}{\,:}&\staS{\vs{Z}} = \staS{X} \land \phi,\\
        \bm{t}{:}&\staT{\vs{Z}} = \staT{X}.
    \end{cases}
    \label{eq:operator3}\\
    \vs{Z} \vDash X + \delta 
    &\iff \bm{f} : 
    \begin{cases}
        \bm{s}{\,:}&\staS{\vs{Z}} = \staS{X},\\
        \bm{t}{:}&\staT{\vs{Z}} = \staT{X} + \delta.
    \end{cases}
    \label{eq:operator4}   
\end{align}

\noindent
It is noteworthy that it is possible to represent the logical \emph{or} operator ($\vee$) similar to~\eqref{eq:operator1} and \eqref{eq:operator3}, whereas~\eqref{eq:operator2} can be used to define other comparison operators. 
Moreover, other mathematical operators like multiplication, division and subtraction can be defined as in~\eqref{eq:operator4}.
Other definitions of operators are legitimate (such as those inspired by Allen's Interval Algebra), as long as they consistently provide a statement in \dom{X} represented with a Boolean state $\bm{s}$, and a time instant $\bm{t}$, as shown in~\eqref{eq:stF}.

\subsection{Fluent Models}
\label{sec:fluent}
Using the formalism briefly drafted in the previous Section, we can design fluent models representing human activity.
As a first example, let us consider an activity consisting of \emph{picking two objects from a cabinet, using them for a while and placing them back in the cabinet}.
In particular, let us consider a smart environment where those objects are associated with sensors able to generate statements $I_4$ and $I_6$ with state $\top$ if the objects are in the cabinet, and $\bot$ otherwise.
Likewise, let us assume that a sensor is attached to the cabinet door $D_7$ with state $\top$ if it is open and $\bot$ otherwise.
We define an activity model $\vs{A}_1$ that uses two aggregated statements, namely \emph{object taken} \vs{T}, and \emph{object released} \vs{R}, such that the activity is considered to be accomplished when those statements hold true after a minimum delay $\delta_1$ between each other. 
The model is as follows: 
\begin{equation}
	{\vs{A}_1 \vDash \big(\big(\vs{T} \land \top\big) + \delta_1\big) \leqslant \big(\vs{R} \land \top\big)}.
	\label{eq:model_a1}
\end{equation} 

\noindent
It must be noted that the model in \eqref{eq:model_a1} is represented as an aggregated statement $\vs{A}_1$, which holds true when the statement $\vs{R}$ becomes true after a $\delta_1$ interval of time since $\vs{T}$ was true.
For the sake of simplification, we denote using $X^\phi$ the result of the \emph{and} operator applied to a statement $X$ and a Boolean state $\phi$ of \eqref{eq:operator3}. 
Therefore, we may define
\begin{equation}
	{\vs{T} \vDash D^\top_7 \leqslant \big(I^\bot_4 \land I^\bot_6\big)},
	\label{eq:model_a1T}
\end{equation}
where the statements $I_4$ and $I_6$ indicate that the objects were not present after the cabinet door was opened.
We also define
\begin{equation}
	{\vs{R} \vDash \big(I^\top_4 \land I^\top_6\big) \leqslant D^\bot_7},
	\label{eq:model_a1R}
\end{equation}
where the statements $I_4$ and $I_6$ indicate that the objects were present before the cabinet door was closed.

More generally, the fluent models that can be designed in \MON are fully compliant with Allen's Interval Algebra, and it is possible to design complex activity recognition models composed of sub-models since $\bm{f}$ is a bijective function, which makes our representation modular and flexible.
Although in this paper we consider examples wherein statements are generated by simple Boolean sensors, it is nonetheless possible to obtain statements as a result of complex computational processes, for example, a classification of hand gestures recognised by neural networks or other methods that take inertial data from a wearable device.
Nevertheless, in order to present the framework clearly, in this paper we will rely on simple models like the one in the example introduced above.
The fluent model related to the example in \eqref{eq:model_a1} is shown in Figure~\ref{fig:model1}, which uses a graphical formalism based on the statement's algebra.

As a second example, let us consider the problem of modelling an activity assumed to be accomplished if the person spends some amount of time in a specific location, \eg \emph{cleaning}.
Similarly to the previous example, let us consider the statement $D^\top_{11}$ when the door of the cabinet containing cleaning tools is open, and $D^\bot_{11}$ otherwise.
We assume that presence sensors generate statements $\vs{L}$ and $\vs{K}$, when the person is located in the living room and in the kitchen, respectively.
For this example, we can design a model that generates a $\top$ aggregated statement, when the door of the cabinet with cleaning tools has been opened, closed and, in the meanwhile, the person spent some time in the two rooms.
For the fluent model of this example, we can define an operator for $\bm{f}$ that counts the number of statements in a given time interval.
To this aim, the \emph{convolution} operator $\circ$ counts statements $X_j \in \chi$ occurring within a time interval $\delta$ starting from the first statement in $\chi$, such as
\begin{equation}
    \label{eq:convolution}
	{\left(\chi^\phi \circ \delta\right) = \left\{X_j : \forall j \in [1,m],\; \staS{X_j} = \phi,\; \staT{X_j} \in [t_0, t_0{+}\delta]\right\}}, 
\end{equation}
where ${t_0 = \min^m_j(\staT{X_j})}$ and $m$ is the total number of statements in $\chi$.
Therefore, the aggregator function $\bm{f}$ using the convolution operator holds true when at least $h$ elements are generated through convolution, \ie
\begin{equation}
    \vs{Z} \vDash \chi^\top \mcirc{h} \delta
    \iff \bm{f} : 
    \begin{cases}
        \bm{s}{\,:}& \staS{\vs{Z}} = \top \iff m \geqslant h,\\
        \bm{t}{:}& \staT{\vs{Z}} = \max^m_j(\staT{X_j}).
    \end{cases}
\end{equation}

\noindent
Given a set of statements representing the location of the person in different rooms, \eg $\chi_{L}$ and $\chi_{K}$, it is possible to define the cleaning activity fluent model as
\begin{equation}
	{\vs{A}_7 \vDash D^\top_{11} \leqslant \big(\big(\chi_{L}^\top \mcirc{h_3} \delta_3\big) \land \big(\chi_{R}^\top \mcirc{h_4} \delta_4\big)\big) \leqslant D^\bot_{11}},
	\label{eq:model_a7}
\end{equation}
wherein ${h_3,h_4\in\dom{N}^+}$ are the minimum number of observations of the person in each room that are assumed to be required for the recognition of the cleaning activity, whereas ${\delta_3,\delta_4\in\dom{N}^+}$ are the time intervals within which the minimum number of observations should take place, and all this should occur before placing back the cleaning tools into the cabinet and closing its door, \ie $D_{11}$ becomes $\bot$.
A graphical representation of this fluent model is shown in Figure~\ref{fig:model7}.

\subsection{Grounding}
\label{sec:grounding}

\PAR{Description Logic Primer}
an OWL-DL ontology represents knowledge using axioms based on \emph{concepts}, \emph{roles} (\ie relationships between concepts or properties associated with a concept), and their \emph{instances}.
In this paper, concepts are denoted by capitalised words, \eg \dltext{ROOM}, roles are denoted by camel-case words, \eg \dltext{connectsWith}, whereas instances are denoted by uppercase letters, \eg \dltext{A} or \dltext{B}. 
Examples of axioms within an ontology include: 
($i$)~\dltext{A:ROOM}, that indicates classification of \dltext{A} as an instance of the concept \dltext{ROOM},
($ii$)~\dltext{ROOM} $\sqsubseteq$ \dltext{LOCATION}, indicating that all the instances of the concept \dltext{ROOM} are also instances of the concept \dltext{LOCATION},
($iii$)~\dltext{(A,B):connectsWith}, denoting a binary relation between the two instances \dltext{A} and \dltext{B}, and finally
($iv$)~\dltext{connectsWith.ROOM}, which defines a concept containing each instance that \dltext{connectsWith} other instances of the concept \dltext{ROOM}.
In the last example we can also include the cardinality restriction used to define \emph{at least}, \emph{at most}, or \emph{exact}, the number of instances (\eg ${\geqslant}2$~\dltext{connectWith.ROOM} $\doteq$ \dltext{CORRIDOR}, or ${=}1$~\dltext{has.DOOR} $\doteq$ \dltext{CABINET}).
Furthermore, axioms can have \emph{conjunctions} $\sqcap$ and \emph{disjunctions} $\sqcup$ between each other.

An ontology can be coupled with an OWL-based reasoner, such as Pellet, which assures consistency among the axioms. 
OWL-based reasoners can also process the Semantic Web Rule Language (SWRL) rules~\cite{horrocks2004swrl}, such that it is possible to generate axioms based on the conjunction of other axioms.
Additionally, an OWL-based reasoner can process a query posed as incomplete axioms (\eg formalised with the SPARQL language), and can provide a solution involving sets of instances, roles or concepts that consistently complete the given incomplete axioms.
A more thorough description of the functionalities of DL-based ontologies can be found in~\cite{baader2003description}. 

\PAR{Statement Grounding on Description Logic}
\label{sec:statementGrounding} 
the network of ontologies is based on the DL formalism and is implemented using the OWL language.
A statement $X\in\dom{X}$ represents an instance \dltext{X} of a concept {$\Omega\sqsubseteq$~\dltext{STATEMENT}}, \ie \dltext{X:$\Omega$}.
Moreover, each statement is associated with a Boolean state by a role, \ie \dltext{(X,\staS{X}):hasState}, and it is also associated with a time instant by a role, \ie \dltext{(X,\staT{X}):hasTime}.
It is noteworthy that we specify only a subset of the roles the \dltext{X} instance must be involved in to logically describe it as an instance of \dltext{STATEMENT}.
However, \dltext{X} might also be described through other DL-based axioms that further specify it as an instance of a generic concept $\Omega$, which can be used to retrieve contextualised statements.
For instance, in the second example proposed in Section~\ref{sec:fluent}, a sensor might generate a set $\chi$ of statements $X_i$ representing the person's location in the environment over time.
In this case, each \dltext{X$_i$} might be classified as an instance of the \dltext{LIVINGROOM} or the \dltext{KITCHEN} concepts, and it would be possible to query an OWL-based reasoner to retrieve the person's location in a contextualised manner, \eg to obtain $\chi_{L}$ or $\chi_{K}$, respectively.

Based on such a contextualised representation of statements, we can develop activity recognition models using a combination of $\bm{f}$ operators formalised through SWRL rules.
In particular, we apply SWRL rules to select statements based on the context, and we use the logic function $\bm{s}$ as well as the algebraic function $\bm{t}$ to define fluent models for activities.
However, SWRL-based models assume the \emph{open-world} assumption and monotonic reasoning entailed by OWL.
On the one hand, the open-world assumption does not allow to compute the maximum value of a set required by~\eqref{eq:operator1}, \eqref{eq:operator2} and \eqref{eq:convolution}, as the reasoner assumes that other unknown elements might exist.
On the other hand, monotonic reasoning forbids the reasoner to generate new statements, but it allows to change the roles of existing statements, \ie change their Boolean state \staS{X} and time instant \staT{X}.
\MON relies on its computational procedures to overcome these limitations through an imperative programming language that the developer can use to design activity models based on a combination of logic-based and imperative-language based paradigms. 
Since it is not possible for an external procedure to access and modify the knowledge in an ontology during the execution of the OWL-based reasoner, the framework synchronises the scheduling of procedures and the execution of the reasoning process.

\section{Scheduling in the Ontology Network}
\label{sec:MON}
\subsection{Ontology Network}
We represent all the knowledge relevant to a certain application domain by spreading it over a network of ontologies.
More formally, we refer to such a network as a tuple ${\bf{N}=\langle\omega,\rho\rangle}$, where  
${\omega\supset\{\mathscr{O}_1, \ldots \mathscr{O}_i, \ldots, \mathscr{O}_d\}}$ is a set of \emph{at least} $d$ nodes encoding ontologies, and 
${\rho\supset\{\mathcal{P}_1, \ldots \mathcal{P}_j, \ldots, \mathcal{P}_b\}}$ is a set of \emph{at least} $b$ computational procedures.
It is to be assumed that the nodes in the network are connected in any combinations among themselves through a set of computational procedures, which retrieve and provide statements from/to a combination of ontologies using high-order functions $\bm{f}$.

The number of ontologies and procedures in a network depends on the application, but the upper ontology ${\mathscr{U}\in\omega}$, and the scheduler ${\mathcal{H}\in\rho}$ (\ie a special type of procedure) are always involved in $\bf{N}$.
On the one hand, $\mathscr{U}$ must contain instances \dlsub{O}{i} and \dlsub{P}{j} that refer to each ontology $\mathscr{O}_i$ and procedure $\mathcal{P}_j$ in the network respectively.
On the other hand, $\mathcal{H}$ gets bootstrapped automatically by \MON, it loads in memory a node $\mathscr{O}_i$ for each ontology \dlsub{O}{i} in $\mathscr{U}$, and initialises the associated OWL-based reasoners. 
Afterwards, \MON continuously observes the state of the $\mathscr{O}_i$ ontologies and, based on the events defining each \dlsub{P}{j} in $\mathscr{U}$, related $\mathcal{P}_j$ procedures are scheduled to pull, aggregate, and push statements for evaluating activity models.

\subsection{The Upper Ontology}
The upper ontology contains definitions pertaining to domain ontologies, procedures, and events.

In the network $\bf{N}$, $\omega$ is the set of nodes that are initialised while bootstrapping an instance of \MON.
Each node $\mathscr{O}_i$ is represented with an instance \dlsub{O}{i} of the \dltext{ONTOLOGY} concept, that is defined in the upper ontology $\mathscr{U}$ as
\begin{equation}
	\label{eq:ontology}
    \begin{aligned}	
	    {=}1\,\textrm{\dltext{represents}}\textrm{\dltext{.IRI}} &\sqcap {\geqslant}1\,\textrm{\dltext{checkedBy}}\textrm{\dltext{.OWLREASONER}}\\
	    &{\doteq}\,\textrm{\dltext{ONTOLOGY}}.
    \end{aligned}
\end{equation}

\noindent
The instance \dltext{\dlsub{O}{i}:ONTOLOGY} is defined by the \dltext{represents} role, which relates \dlsub{O}{i} to the Internationalised Resource Identifier (IRI) of the ontology file containing DL axioms that define a context.
Furthermore, \eqref{eq:ontology} holds knowledge about a reasoner that processes the ontology based on OWL specifications for consistency checking, \eg \dltext{PELLET}$\,\sqsubseteq\,$\dltext{OWLREASONER}.

As discussed above, $\rho$ in $\bf{N}$ is a set of procedures directly defined by \emph{implementations}, which indicate how procedures themselves get computed. 
We simply refer to an implementation as an algorithm developed using some imperative language, which is associated with an activation event triggered on the basis of the context.
Each procedure $\mathcal{P}_j$ exists in the network and it could be scheduled if a relative instance \dlsub{P}{j} is classified as an instance of the \dltext{PROCEDURE} concept, which is defined as follows in the upper ontology $\mathscr{U}$,
\begin{equation}
    \label{eq:procedure}
    \begin{aligned}
        {=}1\,\textrm{\dltext{implements}}\textrm{\dltext{.IRI}}&\sqcap{\geqslant}1\,\textrm{\dltext{requires}}\textrm{\dltext{.EVENT}}\\ 
	    &\doteq\textrm{\dltext{PROCEDURE}}.
    \end{aligned}
\end{equation}
In \eqref{eq:procedure}, the \dltext{implements} role associates an instance \dlsub{P}{j} with an identifier pointing to an algorithm, \ie the implementation of the procedure $\mathcal{P}_j$.
The \dltext{requires} role, instead, associates to \dlsub{P}{j} one or more instances \dltext{E} of the \dltext{EVENT} concept that the scheduler must check before activating the $\mathcal{P}_j$ procedure.

Events are defined in the upper ontology $\mathscr{U}$, and are associated with each \dltext{PROCEDURE} as shown in \eqref{eq:procedure}.
Events represent situations when a statement $X$, within an ontology in $\omega$, assumes a given state $\staS{X}$.
As a matter of fact, in $\mathscr{U}$, each instance \dltext{E:EVENT} is defined as a collection of conditions
\begin{equation}
	\label{eq:event}
	{\geqslant}1\,\textrm{\dltext{observes.CONDITION}}\,{\doteq}\,\textrm{\dltext{EVENT}}.
\end{equation}    
Each condition is in turn defined in $\mathscr{U}$ as an instance \dltext{C} of the concept \dltext{CONDITION}, defined as
\begin{equation}
	\label{eq:condition}
	\begin{aligned}
	    {=}1\,\textrm{\dltext{checks.STATEMENT}}   ~\sqcap~  &{=}1\,\textrm{\dltext{in.ONTOLOGY}}\,\sqcap \\
	    {=}1\,\textrm{\dltext{hasTarget.BOOLEAN}}  ~\sqcap~  &{=}1\,\textrm{\dltext{outcome.BOOLEAN}}\,\sqcap \\                
	    {=}1\,\textrm{\dltext{rate.HZ}}            ~\doteq~  &\textrm{\dltext{CONDITION}}.    
	\end{aligned}
\end{equation}
In \eqref{eq:condition}, the \dltext{checks} role specifies a \dltext{STATEMENT} $X$ to be evaluated \dltext{in} the \dltext{ONTOLOGY} $\mathscr{O}_i$, \eg \dltext{(C,X):checks} where \dltext{X:STATEMENT} and \dltext{(C,\dlsub{O}{i}):in} where \dltext{\dlsub{O}{i}:ONTOLOGY}.
A \dltext{C:CONDITION} would have a $\top$ \dltext{outcome} if and only if the value specified through the \dltext{hasTarget} role is equivalent to the state \staS{X} at certain instants of time, which are based on a \dltext{rate} specified in Hz.

\subsection{Scheduling and Network Management}
For each condition \dltext{C} available in the upper ontology $\mathscr{U}$, the system evaluates one or more logic rules at a given \dltext{rate} to classify \dltext{C} as one of the two disjoint concepts  
\{\dltext{$\top$CONDITION, $\bot$CONDITION}\}\dltext{$\,\sqsubseteq\,$CONDITION},  
depending on whether \dltext{C}'s Boolean state outcome is $\top$ or $\bot$, respectively.
Similarly, we use logic rules to define if an event \dltext{E} is an instance of 
{\dltext{$\top$EVENT}\dltext{$\,\sqsubseteq\,$EVENT}}. 
The event \dltext{E} is related to the condition \dltext{C} via the role \dltext{(\dltext{E},\dltext{C}):observes}. 
Hence, every time \dltext{C} has a change in its \dltext{outcome}, the system applies the rules to classify \dltext{E:$\top$EVENT} when \dltext{E} is related to \dltext{\dltext{C}:$\top$CONDITION} only.
When an instance \dltext{E} is classified in the \dltext{$\top$EVENT} concept, we assume it to be satisfied and, if it holds $\top$ that \dltext{(\dlsub{P}{j},E):requires}, then the scheduler runs the algorithm that such a \dlsub{P}{j} \dltext{implements}. 
It is noteworthy that, through appropriate combinations of conditions, we can specify logical \emph{and} operators among checked outcomes, whereas via a combination of events we can specify logical \emph{or} operators for triggering a procedure.

While a procedure performs computation to aggregate statements, conditions and events semantically identify the context in which such a procedure should be performed. 
In particular, conditions are evaluated through simple computations, \eg check the state of a given statement. 
Hence, the evaluation of an event (\ie a Boolean expression of conditions) is, in qualitative terms, \emph{much simpler} than the execution of a procedure.
As a consequence, the overall reasoning complexity is expected to decrease with respect to the case where the corresponding procedures and contextualised representations are evaluated frequently in a single ontology.

\MON relies on a generic definition of a statement, which can be used not only to represent the knowledge required by activity recognition models via procedures, but also to define the events activating those procedures. 
Therefore, all the ontologies in $\omega$ depend on the upper ontology $\mathscr{U}$.
In particular, on the definition of the \dltext{STATEMENT} concepts (Section~\ref{sec:statementGrounding}), which is shared among all the $\mathscr{O}_i$ ontologies.
In addition, also the procedures in $\rho$ depend on $\mathscr{U}$ since it specifies context-relevant algorithms, the input and output statements they concern, and activation events.
Also, procedures can depend on other ontologies in $\omega$ because their algorithm may require access to certain context-specific statements at runtime. 

An ontology network developed with \MON undergoes two temporal phases, namely bootstrapping and running.
The first phase involves the scheduler activation, which retrieves concept instances represented in the upper ontology $\mathscr{U}$, and generates three maps (shown in Figure~\ref{fig:setup}).
The first map contains all the ontologies $\mathscr{O}_i$, which are initialised in accordance with \eqref{eq:ontology}.
The second map specifies the algorithm that each procedure $\mathcal{P}_j$ \dltext{implements} and their associated \dltext{EVENT} instances \eqref{eq:procedure}.
The third map spans out a \emph{periodic task} $\mathcal{C}_k$ with a specified \dltext{rate} for each \dltext{C:CONDITION}, which is initialised as $\bot$ in the upper ontology.
Such a task is used to schedule procedures based on the outcome of each event, which is determined on the basis of all the \dltext{C} conditions that each event \dltext{observes} \eqref{eq:event}.
In the second phase, the scheduler waits for notifications of condition changes from the periodic tasks.
Based on these notifications, the scheduler eventually classifies some events as \dltext{$\top$EVENT}, and runs the relative procedures.
When a procedure related to \dltext{E} is scheduled, \dltext{E} will not be an instance of \dltext{$\top$EVENT} anymore until consistent conditions re-occur.
Finally, when $\mathcal{P}_j$ is active, the scheduler provides it with the map containing the ontology $\mathscr{O}_i$, and $\mathcal{P}_j$ begins to, pull, aggregate and push statements among ontologies in $\omega$ for evaluating activity models in a synchronised manner. 

\begin{figure}%
    \vspace{.55em}
    \centering
    \includegraphics[width=.906\textwidth]{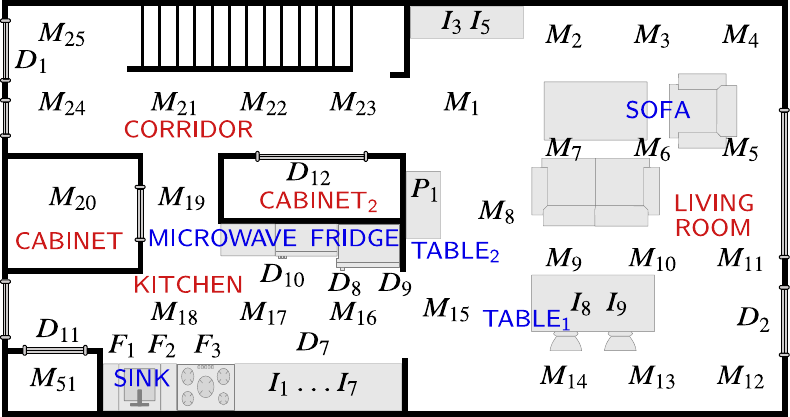}
    \caption{%
        The sensors of a smart home, \ie motion detectors $M_i$, item-presence sensors $I_i$, flow sensors $F_i$, door sensors $D_i$, and phone $P_i$.
        Sensors are contextualised in a topological map concerning areas and furniture.
    }%
    \label{fig:home}%
\end{figure}

\section{Experimental Evaluation}
\label{sec:experiment}

\subsection{A Reference Scenario} 
\label{sec:scenario}

We evaluate \MON using the well-known CASAS dataset from Washington State University\footnote{Available at \url{ailab.wsu.edu/casas/datasets/adlinterweave.zip}}, as it is customary in ambient assisted living research.
The dataset is presented in~\cite{casasDataset}, and it contains data that was collected in an apartment with distributed sensors as shown in Figure~\ref{fig:home}.
When a sensor generates Boolean information at some specific time instants, then statements are produced in the form of \eqref{eq:stF}.
The following sensors are present and considered in this scenario,
($i$)~\emph{motion sensors} generate $\bm{s}(M_i)=\top$ when the presence of a person is detected by the related sensor, 
($ii$)~\emph{item presence sensors} generate $\bm{s}(I_i)=\bot$ when the related sensor detects that the item is absent,
($iii$)~\emph{flow detectors} generate $\bm{s}(F_i)=\top$ when water or gas is flowing,
($iv$)~\emph{door state detectors} generate $\bm{s}(D_i)=\top$ when the related door is open, and
($v$)~\emph{phone usage detector} generates $\bm{s}(P_i)=\top$ when the phone is being used.

Data have been acquired when residents in the apartment were performing a set of activities in an unscripted manner, and therefore data contain interruptions and concurrency in their execution.
In total, 19 volunteers performed the following activities, which are enumerated as
($\alpha_1$)~filling the medication dispenser,
($\alpha_2$)~watching a DVD,
($\alpha_3$)~watering plants,
($\alpha_4$)~conversing on the phone,
($\alpha_5$)~writing a card,
($\alpha_6$)~preparing a meal,
($\alpha_7$)~cleaning the apartment, and
($\alpha_8$)~selecting an outfit from the wardrobe.

\begin{figure}
    \vspace{.55em}
    \centering
    \includegraphics[width=\textwidth]{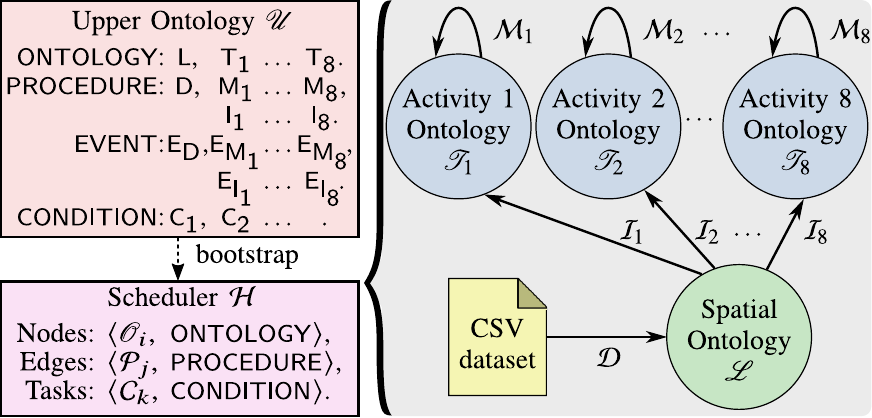}
    \caption{%
        The ontology network presented in Section~\ref{sec:experiment} and developed with \MON for a specific scenario.
        On the right-hand side the network, which is bootstrapped fom the upper ontology, shown on the left-hand side.
        Appendix~\ref{apx:network} details the knowledge in the ontologies (\ie nodes) and their interface through computational procedures (\ie edges) by the means of conditions and events.
    }%
    \label{fig:setup}%
\end{figure}

\subsection{Evaluation Methodology}
\label{sec:eval}

Based on the timestamps available in the dataset, we simulate the streaming of data encoded as statements. 
The statements are used in an ontology network developed for the scenario described above and detailed in the next Section.
We develop the network to show the design supported by \MON and to assess its intelligibility and computational load.

The ontology network is designed to reduce the complexity of each context, \ie ontology. 
The results of such a design are discussed from three different perspectives. 
($i$)~In Section~\ref{sec:comp}, we observe the magnitude of reduction in the computational load. 
The computational load is a focus because logic-based reasoning is exponentially complex with respect to the amount of knowledge encoded in an ontology.
($ii$)~In Section~\ref{sec:expInt}, we discuss the intelligibility level provided by \MON through the examples introduced in Section~\ref{sec:modelImpl}.
($iii$)~In Section~\ref{sec:recRate}, we analyse the differences between human activity recognition performance within an ontology network and approaches that exploit a unique paradigm and/or data representation \emph{corpus}. The main difference being that an ontology network allows to evaluate each activity in a purposely designed, specialised context.

With this experimental setup, we stress again that we do not want to compare our activity recognition models with other techniques validated in the literature, \eg upper ontologies for ADL~\cite{salguero2019ontology} or MLN~\cite{gayathri2017ontoML}, but rather we want to emphasise the unique traits of the architectural aspects of \MON. 
At the same time, we do not aim to investigate the recognition performance for specific activities, \eg meal preparation.
Instead, we want to evaluate an ontology network in terms of its modularity, flexibility and intelligibility while it manages concurrent and interconnected contexts for recognising ADL.

Since \MON enables an iterative development of human activity models, we showcase its modularity using a simple and engineered implementation of such models.
In future \MON iterations, we aim at comparing the recognition performance for specific activities and validating activity models.
Such a validation phase might involve the adoption of computational procedures that are different from the ones used in this paper.
Furthermore, the validation phase may involve semantic representations -- for describing ADL in ontologies -- different from the spatial and temporal representations we propose here.
Nonetheless, \MON allows to flexibly integrate different procedures and ontologies since the only requirement is the proper use of \emph{statements} (to represent data) and \emph{events} (to trigger computation), as discussed in Sections~\ref{sec:overview}--\ref{sec:MON}.

\subsection{The Implementation of an Ontology Network}
\label{sec:impl}
To model the scenario outlined above, we configure the network as shown in Figure~\ref{fig:setup} through a specific configuration of the upper ontology $\mathscr{U}$.
In particular, we configure $\mathscr{U}$ with an instance \dltext{L:ONTOLOGY} representing an ontology that describes a topological map as shown in Figure~\ref{fig:home}.
We configure $\mathscr{U}$ with instances \dltext{T$_a$:ONTOLOGY}, with ${a = 1, \ldots, 8}$, each representing an ontology describing statements over time, \ie based on Allen's Algebra, specifically for each activity $\alpha_1, \ldots, \alpha_8$.
Consequently, we define in $\mathscr{U}$ eight \dltext{M$_a$:PROCEDURE} instances, which implement computational procedures evaluating a fluent model for each $a$-th activity.
Each \dltext{M$_a$} is coupled with another instance \dltext{I$_a$:PROCEDURE}, which imports statements from \dltext{L} to \dltext{T$_a$} when particular events occur.
In this scenario, we also define in $\mathscr{U}$ an instance \dltext{D:PROCEDURE}, which reads the datasets and creates statements in the spatial ontology \dltext{L}, thus simulating an event-based data stream as it would be generated by sensors.

\begin{figure*}
    \vspace{.55em}
    \centering%
    \footnotesize
    \subfloat[\vs{A_1}]{%
        \centering%
        \includegraphics[width=.232\textwidth]{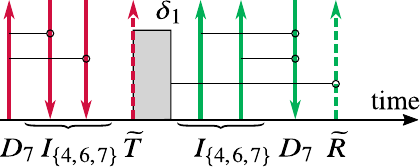}
        \label{fig:model1}%
    }%
    \quad%
    \subfloat[\vs{A_2}]{%
        \centering%
        \includegraphics[width=.232\textwidth]{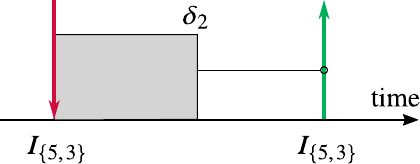}
        \label{fig:model2}%
    }%
    \quad%
    \subfloat[\vs{A_3}]{%
        \centering%
        \includegraphics[width=.232\textwidth]{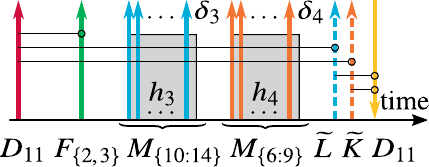}
        \label{fig:model3}%
    }%
    \quad%
    \subfloat[\vs{A_4}]{%
        \centering%
        \includegraphics[width=.232\textwidth]{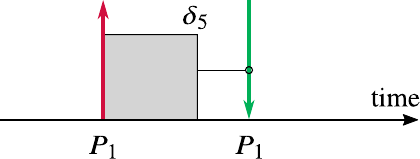}
        \label{fig:model4}%
    }%
    
    \subfloat[\vs{A_5}]{%
        \centering%
        \includegraphics[width=.232\textwidth]{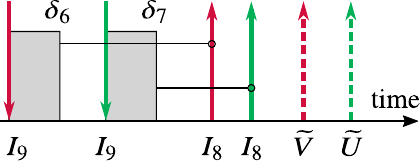}
        \label{fig:model5}%
    }%
    \quad%
    \subfloat[\vs{A_6}]{%
        \centering%
        \includegraphics[width=.232\textwidth]{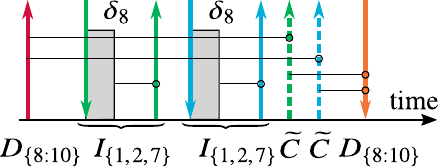}
        \label{fig:model6}%
    }%
    \quad%
    \subfloat[\vs{A_7}]{%
        \centering%
        \includegraphics[width=.232\textwidth]{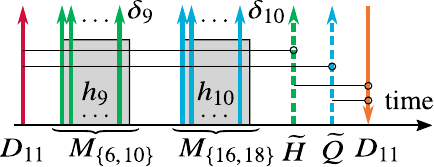}
        \label{fig:model7}%
    }%
    \quad%
    \subfloat[\vs{A_8}]{%
        \centering%
        \includegraphics[width=.232\textwidth]{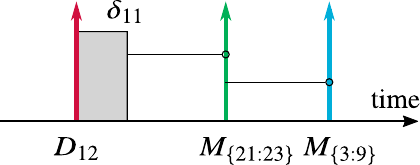}
        \label{fig:model8}%
    }%
    \caption{%
        A graphical representation of the fluent models to classify the activities involved in the scenario presented in Section~\ref{sec:scenario}.
        This representation is described in Section~\ref{sec:impl}, based on the formalisation proposed in Section~\ref{sec:fluent}, and implemented through SWRL rule as detailed in Appendix~\ref{apx:models}.
    }%
    \label{fig:models}
\end{figure*}

During bootstrapping, the scheduler generates a network made with the ontologies $\mathscr{L}$ and $\mathscr{T}_a$, respectively specified by the instances \dltext{L} and \dltext{T$_a$} in $\mathscr{U}$. 
Therefore, the set of ontologies is 
${\omega=\{\mathscr{U},\mathscr{L},\mathscr{T}_1,\mathscr{T}_2,\ldots,\mathscr{T}_8\}}$, 
whereas the set of procedures is 
$\rho=\{\mathcal{H},\mathcal{D},\mathcal{I}_1,\mathcal{I}_2,\ldots,\mathcal{I}_8,\mathcal{M}_1,\mathcal{M}_2,\ldots,\mathcal{M}_8\}$,
which includes the scheduler $\mathcal{H}$ itself, and the procedures associated with the instances \dltext{D}, \dltext{I$_a$} and \dltext{M$_a$} respectively.
The procedures also specify events, which are associated with $\mathcal{C}_k$ conditions \eqref{eq:procedure}, \eqref{eq:event}, \eqref{eq:condition}.

The following paragraphs introduce the spatial ontology $\mathscr{L}$, the computational procedures $\mathcal{M}_a$ (which implements fluent models), the activity ontologies $\mathscr{T}_a$, and the scheduling events.
For in-depth detail, Figure~\ref{apx:tab:onto} and Appendix~\ref{sec:apx:onto} focus on $\mathscr{L}$ and $\mathscr{T}_a$, Appendix~\ref{sec:apx:proc} report scheduling events and the procedures in $\rho$, whereas Appendix~\ref{apx:models} discusses fluent model implementations in $\mathcal{M}_a$. 

\PAR{The spatial ontology} 
$\mathscr{L}$ contextualises spatial information, and it contains only the most up to date statements generated by each sensor in the smart home since it overwrites previous values for those statements.
Therefore, the number of axioms does not grow over time.
This is a crucial feature since it allows to synchronise the procedures in the network, as it is possible to identify \emph{a priori} a reasoning time limit for the spatial classification of statements.
In $\mathscr{L}$ are defined such spatial roles as \dltext{isIn} and \dltext{isNearTo}, which stand for sensory \dltext{STATEMENTS} related to concepts referring to \emph{locations} (\eg \dltext{KITCHEN}) and \emph{furniture} (\eg \dltext{TABLE}).
When a sensor generates new data, $\mathcal{D}$ creates a new statement in $\mathscr{L}$, and it updates the corresponding OWL reasoner, which thus infers the person's location and the furniture that he (she) might be using. 
As we will see in later paragraphs, each computational procedure $\mathcal{I}_a$ is scheduled based on the locations and furniture that we define to be relevant for the $a$-th activity.

\PAR{Fluent model implementations}
\label{sec:modelImpl}
fluent models are based on the formalism presented in Section~\ref{sec:statement}, which is also graphically shown in Figure~\ref{fig:models}, where statements are depicted as vertical arrows pointing upwards to indicate a $\top$ state and downwards to indicate $\bot$ state. 
The Figure shows the statements with respect to a relative and qualitative temporal axis, which defines the restriction making a fluent model generate an aggregated statement $\bm{s}(\vs{A_a})=\top$, thus indicating that the $a$-th activity has been performed at time $\bm{t}(\vs{A_a})$.
Solid arrows represent raw sensor statements, whereas dashed arrows indicate aggregated statements, \ie the results of the aggregation of other statements from a procedure.
Colours indicate sub-models, \ie statements in the sub-models respect the precedence operators in $\bm{f}$.
Figure~\ref{fig:models} shows time intervals of a specified span $\delta$ with grey boxes and denotes with $h$ the minimum number of statements expected to be in an interval.
Through horizontal lines, the Figure depicts restrictions over time among the statements that must hold for satisfying a model, where the cycled statement is required to be always after the other.

As an example, Figure~\ref{fig:model2} represents the model
\begin{equation} 
    \label{eq:mm2}
    \vs{A}_2 \vDash \big(I^\bot_{\{5,3\}} + \delta_2\big) \leqslant I^\top_{\{5,3\}},
\end{equation}
which is satisfied when the item observed by sensors $I_5$ or $I_3$ has a $\bot$ state and, after a $\delta_2$ span of time, it becomes $\top$, which means that the objects needed for watching a DVD have been used \emph{for a while}.
Figure~\ref{fig:model4} shows the same model applied to the $P_1$ sensor for identifying the activity of conversing using the phone, and Figure~\ref{fig:model8} shows a similar model for outfit selection.
In this case, we assume that an outfit is chosen when the wardrobe door is open, and after a $\delta_{11}$ time interval, the person is still in the corridor, \ie the region associated with sensors $M_{21}$, $M_{22}$ and $M_{23}$.
We assume that the person places the selected outfit on the sofa, therefore we specify that in order to perform the activity the person should visit the region monitored by $M_3$ up to $M_9$. 
In Section~\ref{sec:statement}, as examples, we discussed the models \vs{A_1} and \vs{A_7}, which are related to medical dispenser filling and the cleaning activity, respectively.
Similarly to the case of \vs{A_1}, Figure~\ref{fig:model5} represents \vs{A_5}, which is related to the activity of writing a card.
\vs{A_5} involves using two objects related to sensors $I_8$ or $I_9$ from some time, \ie
\begin{equation}
    \label{eq:mm5}
    \vs{A_5} \vDash \vs{U} \land \vs{V} \vDash \big(\big(I^\bot_8+\delta_6\big)\leqslant I^\top_8\big) \land \big(\big(I^\bot_9+\delta_7\big)\leqslant I^\top_9\big).
\end{equation}
Furthermore, \vs{A_3} is recognised if the person remains in some particular location based on the convolution operator, \ie
\begin{align}
    \label{eq:mm3}
    &\vs{A_3} \vDash D^\top_{11} \leqslant F^\top_{\{2,3\}} \leqslant \big(\vs{G} \land \vs{E}\big) \leqslant D^\bot_{11},\\
    \text{where}\;\;\;&
    \vs{G}   \vDash   M_{\{6:9\}}\mcirc{h_3}\delta_3
    \;\;\;\text{and}\;\;\;
    \vs{E}   \vDash   M_{\{10:14\}}\mcirc{h_4}\delta_4. \notag
\end{align}
\vs{A_3} represents the activity of watering the plants, which can be graphically shown to domain experts as in Figure~\ref{fig:model3} or be expressed in natural language as
\expl{$D_{11}$ was opened and $F_{\{2,3\}}$ was used, then the person stayed in $M_{\{6:9\}}$ for $\delta_3$ time units, and in $M_{\{10:14\}}$ for $\delta_4$ time units, then $D_{11}$ got closed,}
where $D_{11}$, $F_{\{2,3\}}$, $M_{\{6:9\}}$ and $M_{\{10:14\}}$ are statements in $\mathscr{L}$ and $\mathscr{T}_3$, which are based on \emph{a priori} knowledge about the sensors.
In $\mathscr{L}$, they are contextualised in terms of locations and furniture, \ie respectively \dltext{CABINET}, \dltext{SINK}, \dltext{PLANT1} and \dltext{PLANT2}, which are involved while watering the plants.
In $\mathscr{T}_3$, sensor statements are contextualised with the prior knowledge required only to evaluate \vs{A_3} \eqref{eq:mm3}, \ie sensor types as \{\dltext{DOOR,FLOW,WATERED}\} and time intervals $\delta_3$ and $\delta_4$.

For all the eight activities, Appendix~\ref{apx:models} highlights intelligibility of the prior knowledge and of the statement's algebra, as it is presented above for $\vs{A_3}$.
It is noteworthy that \MON contextualises statements based on a general-purpose hierarchy of DL concepts, and it is possible to further specify their semantics within a context, \eg that ``a person is near \dltext{PLANT1}'', or that ``a plant has been \dltext{WATERED}'', based on locations or time, respectively.
As long as an ontology remains small, a detailed classification of statements is intelligible and simple; also, the reasoning complexity is limited.

\PAR{Activity ontologies}
$\mathscr{T}_a$ are bootstrapped via the instances of the \dltext{ONTOLOGY} concept defined in the upper ontology $\mathscr{U}$.
Each activity ontology holds knowledge that a procedure $\mathcal{M}_a$ evaluates to generate an aggregated statement, \eg $\vs{A_a}$, which represents the fact that $\alpha_a$ has been carried out at a certain time instant $\bm{t}(\vs{A_a})$. 
As opposed to $\mathscr{L}$, which represents only a spatial context, for an ontology $\mathscr{T}_a$ each procedure $\mathcal{I}_a$ introduces temporal statements without overwriting the old statements.
In particular, the $\mathcal{I}_a$ procedure \emph{observes} the spatial ontology $\mathscr{L}$ and \emph{moves} part of the statements related to the $a$-th activity into $\mathscr{T}_a$.
This is designed to strictly represent in $\mathscr{T}_a$ only the knowledge required for the fluent model that $\mathcal{M}_a$ defines.
For simplicity, we remove all the statements from $\mathscr{T}_a$ when the activity is recognised, \ie when $\bm{s}(\vs{A_a}) = \top$ exists in $\mathscr{T}_a$.
However, this approach might not be suitable for less controlled scenarios, and this opens challenging issues.

Fluent models are implemented using the SWRL rules.
Therefore, procedures $\mathcal{M}_a$ can assert \vs{A_a} statements in the ontology by updating the OWL reasoner associated with $\mathscr{T}_a$.
As discussed in Section~\ref{sec:grounding}, we cannot compute the minimum or maximum value of a set using an SWRL rule.
To compute the logic operations \eqref{eq:operator1} and \eqref{eq:operator2}, we could retrieve the maximum timestamp and satisfy the statement's algebra since models have an ending event, \eg a closed door, or a visited place.
On the contrary, performing convolution operations \eqref{eq:convolution}, \eg used for activities $\alpha_3$ and $\alpha_7$, is an issue.
In order to overcome this limitation, we delegated to the $\mathcal{M}_3$ and $\mathcal{M}_7$ procedures the evaluation of the convolution of statements.
The result of the convolution is another statement that $\mathcal{M}_3$ and $\mathcal{M}_7$ store in the ontologies $\mathscr{T}_3$ and $\mathscr{T}_7$, respectively,
Then, $\mathcal{M}_3$ and $\mathcal{M}_7$ can update the reasoner of $\mathscr{T}_3$ and $\mathscr{T}_7$ to evaluate the related activity models based on SWRL rules.
This highlights that \MON can be used to create hybrid approaches, since the computational procedures can exploit a combination of imperative programming languages and OWL.

\PAR{Events}
\label{sec:eventImpl}
they are bootstrapped from the upper ontology into a set of tasks that periodically evaluate the outcome of each condition $\mathcal{C}_k$.
The upper ontology describes the $\mathcal{D}$ procedure using an instance represented with a role \dltext{(D,E$_\textrm{\dltext{D}}$):requires}, where \dltext{E$_{\textrm{\dltext{D}}}$} is an event occurring once after the bootstrap phase.

The events activating a procedure $\mathcal{I}_a$ depend on the knowledge required by the $a$-th activity model.
For the $\vs{A_1}$ model, $\mathcal{I}_1$ maps statements from $\mathscr{L}$ to $\mathscr{T}_1$ when the person is in the kitchen. 
For this to happen, the event \dltext{E$_\textrm{\dltext{I$_1$}}$} is triggered based on the spatial ontology $\mathscr{L}$.
In other words, $\mathcal{I}_1$ depends only on an event composed of a \dltext{C$_1$:CONDITION}, which has a $\top$ outcome if a number of sensors placed in the living room have a $\top$ state.
When $\mathcal{I}_1$ is activated, it retrieves statements $D_7$, $I_4$, $I_6$, $I_7$ from $\mathscr{L}$, and maps them into $\mathscr{T}_1$, whose purpose is to contextualise knowledge over time to eventually generate a $\top$ outcome for the aggregated statement \vs{A_1} (Figure~\ref{fig:model1}).
Remarkably, we used the same modular approach for all the eight activities in the dataset, where the set of statements retrieved from $\mathscr{L}$ and mapped into $\mathscr{T}_a$ change with respect to the knowledge strictly required to evaluate the $a$-th activity model (Figure~\ref{fig:models}).

We rely on similar events for all the $\mathcal{I}_a$ procedures but, since each activity model requires different contextualised statements, their spatial semantics vary.
In particular, the scheduler activates the procedure $\mathcal{I}_a$ based on a \dltext{E$_\textrm{\dltext{I$_a$}}$:EVENT} such that there is a $\top$ outcome when the person is
(\dltext{E$_\textrm{\dltext{I$_2$}}$}) in the \dltext{LIVINGROOM}, 
(\dltext{E$_\textrm{\dltext{I$_3$}}$}) near to \dltext{CABINET$_1$}, or to the \dltext{SINK}, or in the \dltext{LIVINGROOM},
(\dltext{E$_\textrm{\dltext{I$_4$}}$}) near to \dltext{TABLE$_2$},
(\dltext{E$_\textrm{\dltext{I$_5$}}$}) near to \dltext{TABLE$_1$}, 
(\dltext{E$_\textrm{\dltext{I$_6$}}$}) in the \dltext{KITCHEN},
(\dltext{E$_\textrm{\dltext{I$_7$}}$}) in the \dltext{LIVINGROOM}, or in the \dltext{KITCHEN}, and
(\dltext{E$_\textrm{\dltext{I$_8$}}$}) in the \dltext{CORRIDOR}, or near the \dltext{SOFA} or \dltext{TABLE$_1$}, where the person should leave the selected outfit (Figure~\ref{fig:home}).
Appendix~\ref{apx:network} provides more details about the scheduling events and the statements required and produced by the procedures, \ie their interface with the ontology.

The procedure $\mathcal{M}_a$ is based on synthetic events that $\mathcal{I}_a$ generates after importing new statements from $\mathscr{L}$ to $\mathscr{T}_a$.
In particular, $\mathcal{I}_a$ generates that event by storing in $\mathcal{T}_a$ a specific $\top$ statement%
\footnote{The statement is expressed as \dltext{N} in Table~\ref{apx:tab:proc} and discussed in Appendix~\ref{apx:network}.},
which is reset to $\bot$ by $\mathcal{M}_a$ at the beginning of its computation.
This and other similar approaches can always be used to synchronise two procedures in a network.

\section{Results}
\label{sec:results}

\subsection{Computational Load and Distribution}
\label{sec:comp}

\begin{figure}
    \vspace{.55em}
	\centering
	\includegraphics[width=.99\textwidth]{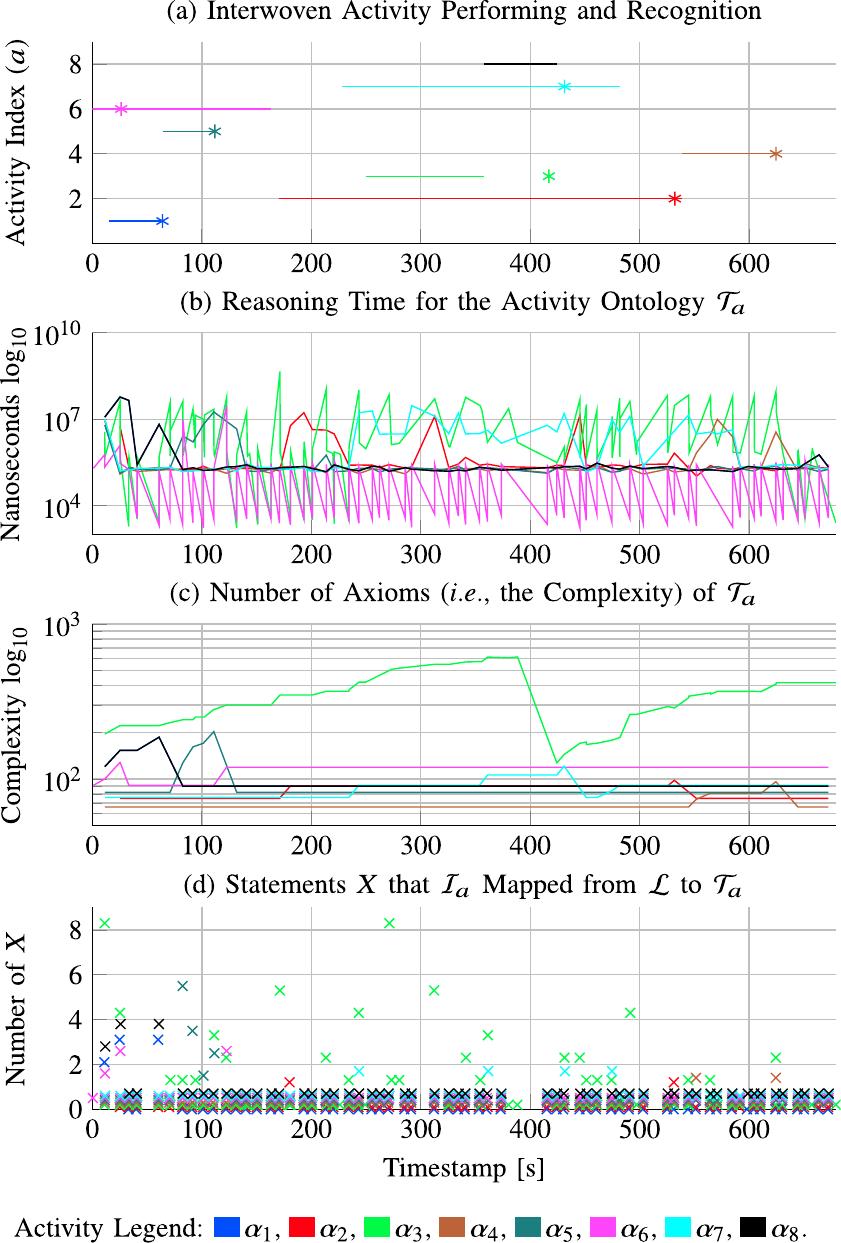}
	\caption{%
	    The behaviour of the ontology network shown in Figure~\ref{fig:setup}. 
	    Graph (a) shows the experiment and the associated recognitions. 
	    Graphs (b) and (c) show the reasoning performance. Graph (d) shows the events in the network and the statements $X$ propagated from the spatial ontology $\mathscr{L}$ to the relevant activity ontology $\mathscr{T}_a$.
	 }
     \label{fig:subPlot}
\end{figure}

Figure~\ref{fig:subPlot} shows the behaviour of a network of ontologies configured as presented in Section~\ref{sec:experiment}, and running on an Intel i5 $2.53$ GHz powered workstation with 4 Gb of memory.
The Figure shows four graphs with the same timestamp on the $x$ axis and a colour map whereby each colour indicates an activity.
The $y$ axis of the first plot represents the activity labels given in the dataset with a horizontal line, \ie the time spent in performing an activity, and the \textasteriskcentered~symbol indicates the time instant when \MON notifies that the activity has been executed, \ie $\bm{t}(\vs{A_a})$ when $\bm{s}(\vs{A_a})=\top$.
The second and the third sub-plots present the reasoning performance for each activity ontology $\mathscr{T}_a$ on a logarithmic scale.
Specifically, the plots show the computation time in nanoseconds, and the ontology complexity (\ie the number of axioms in the ontology), respectively. 
Figure~\ref{fig:subPlot}d shows the number of statements propagated from the spatial ontology $\mathscr{L}$ to $\mathscr{T}_a$ via the importing procedure $\mathcal{I}_a$.

From a comparison of the four sub-plots, we can analyse the reasoning behaviour with respect to the statements contextualised in the spatial ontology $\mathscr{L}$.
The scheduler evaluates statements to detect events and trigger a procedure $\mathcal{I}_a$, which in turn retrieves other statements from $\mathscr{L}$ and adds them to $\mathscr{T}_a$.
The number of mapped statements are shown with a $\times$~point in Figure~\ref{fig:subPlot}d, while the complexity of $\mathscr{T}_a$ is shown in Figure~\ref{fig:subPlot}c. 
Figure~\ref{fig:subPlot}b shows the computation time required to process the statements.
Figure~\ref{fig:subPlot}a shows when the models are satisfied with respect to the activity's ground-truth, from which we can also notice the interruptions occurring in the activities.

\begin{figure}
    \vspace{.55em}
    \centering
    \footnotesize 
    \includegraphics[width=.99\textwidth]{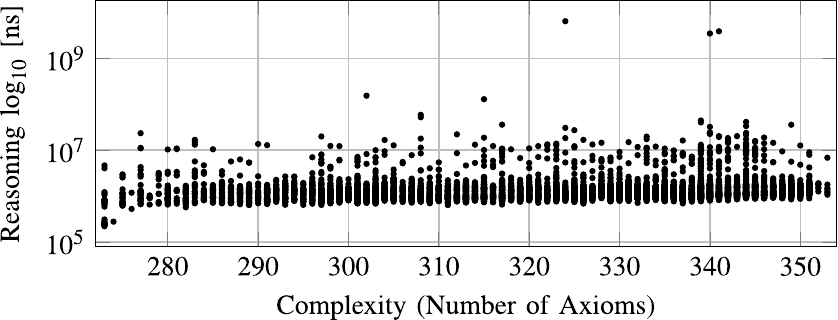}
    \vspace{-.7em}
    \caption{The spatial ontology $\mathscr{L}$ reasoning time against ontology complexity.}%
    \label{fig:placingTime}
\end{figure}

Figure~\ref{fig:subPlot}b shows a frequent pattern as the conditions are evaluated at a rate of $50$ Hz, which drives the event evaluation and, consequently, the scheduling of all the $\mathcal{I}_a$ and $\mathcal{M}_a$ procedures.
Depending on the context over time, not all events generate statements that get evaluated with the procedure $\mathcal{M}_a$, and this leads to the spikes shown in Figure~\ref{fig:subPlot}b.
In particular, the scheduler synchronously activates $\mathcal{M}_a$ when a statement is mapped by $\mathcal{I}_a$.
As a consequence, $\mathcal{M}_a$ calls the reasoner associated with the $\mathscr{T}_a$ ontology, which evaluates activity models $\alpha_a$ based on a temporal representation.

Since we used stream reasoning techniques~\cite{stemingReasoning}, the reasoning complexity depends on an ontology's previous state, and the time required to check whether statements in $\mathscr{T}_a$ satisfy an activity model is dependent on the context over time.
For this reason, the characterisation of the reasoning complexity for an ontology is far from trivial, and we use the number of axioms that it contains as an approximation.
However, this issue is limited to the spatial ontology $\mathscr{L}$, whose reasoner (invoked by the $\mathcal{D}$ procedure) evaluates all the sensor statements over time.
Figure~\ref{fig:placingTime} shows that the reasoning complexity of the spatial ontology is similar to the one characterising activity ontologies, although with a smaller variance.
Also, Figure~\ref{fig:placingTime} shows that the number of axioms used to represent prior spatial knowledge without any sensor statements is $273$, \ie the lower complexity bound of $\mathscr{L}$.
Instead, the upper bound of $353$ axioms shows the maximum complexity of $\mathscr{L}$, and it cannot increase due to the overwriting policy of $\mathcal{D}$.

In Figures~\ref{fig:subPlot}c, we observe an exponential increase of the complexity over time for not well-contextualised activities.
For instance, $\alpha_3$ relies on a model that requires statements from many locations in the apartment, \ie the spatial context related to the model \vs{A_3} (Figure~\ref{fig:models}) is not \emph{well-separated} enough from other activities as far as its definition is concerned.
Instead, for activities characterised by a more distinct context, \eg $\alpha_2$ and $\alpha_4$, the complexity remains limited.
In this experiment, we delete all the statements from $\mathscr{T}_a$ when the $a$-th activity has been recognised. 
The effect of this can be seen by noticing the drop of the curves in Figure~\ref{fig:subPlot}c with respect to the instants when an activity is recognised, as shown in Figure~\ref{fig:subPlot}a.
This statements' cleaning procedure is associated with tackling the issue of continuous reasoning that often leads to an exponential increase in the complexity~\cite{scalmato2013owl,noor2018ontology}.

\begin{table}
    \vspace{.55em}
    \renewcommand{\arraystretch}{1.1}
    \centering
    \floatbox[{\capbeside\thisfloatsetup{capbesideposition={right,center},capbesidewidth=3.5cm}}]{table}[\FBwidth]{
        \caption{%
            Delay in activity recognition, for a simulation that is four times faster than the real data stream, for a total of 19 volunteers; the maximum, and average delay in seconds.
        }%
        \label{tab:delay}
    }{%
        \footnotesize
        \begin{tabular}{c|c|c|c}
                           & delayed    & max [s]   & avg [s]    \\ \hline
			$\alpha_1$     & 0/19       & $-$       & $-$        \\ \hline
			$\alpha_2$     & 4/19       & 9.46      & 4.17       \\ \hline
			$\alpha_3$     & 8/19       & 17.3      & 9.2        \\ \hline
			$\alpha_4$     & 5/19       & 12.1      & 5.83       \\ \hline
			$\alpha_5$     & 0/19       & $-$       & $-$        \\ \hline
			$\alpha_6$     & 1/19       & 4.17      & $-$        \\ \hline
			$\alpha_7$     & 1/19       & 4.6       & $-$        \\ \hline
			$\alpha_8$     & 5/19       & 9.2       & 5.41
	    \end{tabular}
    }%
\end{table}

Due to an increase in the amount of knowledge to process, reasoning becomes exponentially complex, and this can also be noticed by comparing the activities $\alpha_3$ and $\alpha_6$ in Figure~\ref{fig:subPlot}.
In this case, $\alpha_6$ requires less computation time than $\alpha_3$, since the former has a well-defined spatial contextualisation, while the latter occurs in a not well-determined area.
Therefore, having a small and well-contextualised ontology $\mathscr{T}_a$ decreases the computational load, and it is more likely that the related model is processed only when required.
We tested the performance of the network by accelerating the simulation $4\times$ faster than its original speed, and we did not observe notable changes in the activity recognition rate.
Instead, for higher speeds, the recognition performance drastically decreases due to the overwriting policy of $\mathcal{D}$.
The delay in the recognition we measure during the accelerated simulation is shown in Table~\ref{tab:delay}.

We attempted to test an ontology network made of a single node such that it encodes time intervals, spatial knowledge, and all the activity models; as presented in Sections~\ref{sec:impl}.
Our objective was to compare the reduction of computational load between an approach that uses a single ontology and the ontology network we develop.
Unfortunately, we cannot report the comparison because we experienced \emph{out of memory} issues.
In our previous work \cite{yuhasTh}, we used \MON to develop another ontology network designed for a scenario similar to the one presented in Section~\ref{sec:scenario} but involving fewer activities and sensors.
In that work, we observed that the reasoning time scales exponentially when a single ontology becomes more complex, whereas the overall reasoning time scaled linearly with that network.
Although more evaluations are required to fully characterise the complexity of a general ontology network designed with \MON, in this paper we can confirm a reduction of the computational load.
Indeed, with a network of small contexts we could recognise activities with \emph{soft} real-time constrains, whereas if the same knowledge is encoded in a single ontology, we could not process the data stream.

\begin{table}
    \vspace{.55em}
    \footnotesize
    \centering
    \begin{tabular}{@{}c@{~~}|@{~~}c@{~~}|@{~~}c@{~~}|@{~~}c@{~~}|@{~~}c@{~~}|@{~~}c@{~~}|@{~~}c@{~~}|@{~~}c@{~~}|@{~~}c@{~~}|@{~}c@{}}
 & \cite{Begg2006} & \cite{casasDataset} & \cite{Fleury2010} & \cite{nazerfard2013} & \cite{riboni2016multi} & \cite{gayathri2017ontoML} & \cite{matassa2019reasoning} & \cite{salguero2019ontology} & \MON \\ \hline
        $\alpha_1$ & .66  & .66  & .77  & .77  & .85  & .95  & .68 & .89 & .97 \\ \hline
        $\alpha_2$ & .69  & .86  & .73  & .78  & .81  & .97  & .98 & .91 & .92 \\ \hline
        $\alpha_3$ & .68  & .29  & .75  & .79  & .72  & .94  & .30 & .72 & .80 \\ \hline
        $\alpha_4$ & .66  & .59  & .75  & .78  & .72  & .95  & .87 & .34 & .98 \\ \hline
        $\alpha_5$ & .61  & .83  & .72  & .79  & .81  & .98  & .98 & .78 & .89 \\ \hline
        $\alpha_6$ & .65  & .83  & .75  & .79  & .88  & .97  & .74 & .98 & .80 \\ \hline
        $\alpha_7$ & .64  & .88  & .72  & .78  & .57  & .96  & .76 & .84 & .78 \\ \hline   
        $\alpha_8$ & .66  & .67  & .72  & .79  & .88  & .95  & 1.0 & 1.0 & .97  
    \end{tabular}
    \\[.5em]    
    \begin{tabular}{@{}c@{\,}p{.44\textwidth}@{}c@{\,}p{.45\textwidth}@{}}
        \cite{Begg2006}             &  Artificial Neural Network,      &  \cite{casasDataset}         &  Hidden Markov Model,\\ 
        \cite{Fleury2010}           &  Support Vector Machine,         &  \cite{nazerfard2013}        &  Bayesian network, \\ 
        \cite{riboni2016multi}      &  MLN with numerical constraints, &  \cite{gayathri2017ontoML}   &  MLN extended with OWL,\\ 
        \cite{matassa2019reasoning} &  Hybrid OWL-statistical system,  &  \cite{salguero2019ontology} &  OWL Class Expression Learning.
    \end{tabular}
    \vspace{-1.5em}
    \caption{F-measure of activity recognitions shown for the ontology network developed with \MON (Section~\ref{sec:experiment}) and approaches using a single context to represent data.}
    \label{tab:Fmeasure}
\end{table}

\subsection{Explainability and Intelligibility}
\label{sec:expInt}

Section~\ref{sec:modelImpl} presents the fluent models we use in our scenario.
They are based on the formalism presented in Section~\ref{sec:statement} and are graphically visualised in Figure~\ref{fig:models}.
We confirm that a DL-based formalism is intelligible and can be understood by a person for having a contextualised explanation of activities and events.
Appendix~\ref{apx:models} details the implementation of the models based on SWRL rules, and it also provides a possible sentence in natural language that can be produced from the statement's algebra to explain a recognised activity. 

The intelligibility of our models is facilitated as we consider small contexts.
Having small contexts allows domain experts to focus on their field of interest while considering a specific activity.
Thus, \MON is more intelligible than a complex single ontology concerned with all the activity models.

Detecting the preparation of a meal ($\alpha_6$) is trivial if we can assume that a specific sensor generates events that only occur during that activity, \eg the event of taking a soup.
However, the same knowledge could be generated using a more complex algorithm and the domain experts would not be aware of the difference.
For instance, a neural network could process images for recognising which object has been taken from a cabinet.
In this case, although \MON would not be able to explain the gesture made while taking that object, it could nonetheless explain how that object was used with respect to another piece of furniture, \eg the microwave. 

Thanks to the intelligibility of \MON, we could assess the activity ontologies $\mathscr{T}_a$ regarding why an activity was not recognised.
Such an explanation is retrieved by looking at the symbols in $\mathscr{T}_a$ after the data related to a person has been processed.
For instance, we observed that the 5\% of misclassifications of the $\alpha_2$ activity was due to the fact that the DvD was not released in its expected place.
Furthermore, 15\% of unaccomplished recognition of the $\alpha_5$ activity occurred because the person used different items to prepare the meal, while 25\% of classification errors occurred for $\alpha_6$ because the person did not use the items on the table to write a card. 
Finally, in case of $\alpha_7$, 15\% of the cleaning room activity was misclassified because other activities were occurring in the same location and being performed with close interruptions.
This sort of information might be valuable for domain experts during future network development iterations.

\subsection{Activity Recognition Performance}
\label{sec:recRate}

Table~\ref{tab:Fmeasure} shows the F-measure of the activity recognition rates compared among ours and other approaches for the same dataset.
The other techniques are based on ML and probabilistic approaches, as well as on context-based systems exploiting hybrid OWL-statistical reasoning.
As mentioned above, Table~\ref{tab:Fmeasure} does not indicate that the activity models we developed in Section~\ref{sec:modelImpl} are expected to outperform other approaches in a less controlled scenario.
Our aim is to ground the possibility that having multiple ontologies does not jeopardise, in principle, the possibility of obtaining results comparable to other methods.
These techniques approach the problem with a different paradigm, \ie they are inclined towards designing a \emph{single context} to ground all the activity models in the dataset.
Instead, our paradigm involves hierarchical data representations aimed to accommodate each activity model within a purposely-tailored context.

Table~\ref{tab:Fmeasure} shows that, in this scenario, symbolic reasoning based on hierarchical and concurrent contexts can exhibit results comparable with other state of the art approaches.
However, our activity models require an engineering process that is hard to generalise, especially due to the limited size of the dataset. 
Hence, we cannot use the comparison to evaluate the recognition rate of specific activities.
Nevertheless, considering that \MON has been designed to support an iterative development process, and that the network presented in Section~\ref{sec:impl} has been implemented with simple heuristics, Table~\ref{tab:Fmeasure} confirms that managing hierarchical and concurrent contextualisations of data could be effective for human activity recognition.
This is true if suitable \emph{a priori} knowledge is encoded in \MON, which highlights the need for undertaking a development process guided by domain experts. 

\section{Discussion} 
\label{sec:discussion}

Section~\ref{sec:statement} presents a general-purpose formalisation of atomic knowledge, \ie statements.
Based on this representation, we can develop a formal interface between computational procedures and the ontologies of a network.
A computational procedure can implement an arbitrary algorithm as long as it can meet certain requirements, which are discussed in Section~\ref{sec:overview}.
Those constraints specify that a procedure's inputs and outputs are statements, and that its computation is scheduled based on events, which are based on statements' classification.
Since statements are based on the general-purpose OWL formalism, \MON is flexible and does not further limit the design of the computational procedures.

In order to assure intelligibility, computational procedures store their outcomes as contextualised statements based on \emph{a priori} knowledge.
In Section~\ref{sec:fluent} and \ref{sec:modelImpl}, we describe a few models developed for recognising activities in a specific scenario, and we show their graphical representation in Figure~\ref{fig:models}.
With the support of Appendix~\ref{apx:models}, we also emphasize how statements in ontologies are intelligible for a domain expert.

Since OWL reasoning is exponentially complex with respect to the number of axioms in the ontology~\cite{baader2003description}, \MON's design assumes that the \emph{size} of an ontology must be reduced. 
To this aim, \MON's approach is to distribute knowledge over a network of ontologies.
In each node of the network, reasoning is based on, different, contextualised, \emph{a priori} available knowledge, which is meant to address a specific, but partial, aspect of the application context. 
As discussed in Section~\ref{sec:relatedwork}, since each ontology in the network is associated with a dedicated reasoner, we can use in the same architecture data contextualised with a varying trade-off between performance and expressivity level for each activity. 
Instead, approaches employing a single context paradigm need to mitigate the trade-off among all the activities involved in an application.

We observed that using a set of small ontologies has three main benefits. 
($i$)~It can make the architecture intrinsically modular, since ontologies are pairwise independent, and we can use computational procedures to mutually map them.
($ii$)~Having small ontologies reduces the computational load not only because the associated reasoners evaluate small fragments of knowledge, but also because they rely on events to reason when it is required.
During the design phase, if we noticed that represented knowledge was becoming too complex, then we would avoid increasing the complexity of ontologies.
Instead, we would increase the number of ontologies and limit the amount of knowledge encoded in each of them. 
In our experiments, we observed that an ontology network has a manageable computational load and we tested the limits of the network by artificially accelerating the incoming data stream simulation. 
($iii$)~Taken as a representation framework, a network of small ontologies is more intelligible than a unique, complex representation.
A domain expert can limit their modelling efforts on specific activity models, which are well-separated, semantically-defined modules in the network.
As a drawback, \MON requires an extensive phase of knowledge engineering to design all contexts representations and combine them in a network. 

In the implementation of an ontology network (Section~\ref{sec:impl}), we do not consider the challenging issues related to multi-occupancy, and in this scenario \MON assumes that there is a single person in the smart home.
Due to its representative nature, to perform multiple person activity recognition and classification, \MON would require computational procedures evaluating sensory data tagged with a person identifier.
\MON could then be used to instantiate branches in the network of ontologies that are dedicated to each assisted person.
The same modular approach could also be used to test and compare other human activity models designed within an iterative development process.

The current implementation of \MON does not take into account uncertainties, and it could not be robust to errors and missing information in input data.
In order to overcome this limitation, our formalisation is consistent with other OWL reasoners supporting fuzzy~\cite{f2016fuzzya} or probabilistic~\cite{r2016Probabilistica} logic.
Moreover, with an ontology network, we could make the network redundant, \ie using multiple approaches for recognising the same activity, and gain robustness to noise and error by exploiting contextualisation and voting paradigms. 

As mentioned in Section~\ref{sec:relatedwork}, \MON should be used only with event-based sensors.
In order to exploit sensors that provide continuous data streams (\eg inertial data or videos), an ontology network requires computational procedures that generate statements by processing such data.
Along with an algorithm to be computed, there is also the need to define prior knowledge for contextualising its outcomes, \eg the semantic description of a gesture or a postural transition.
Describing such concepts in a logic-based framework might be difficult but, since \MON supports an iterative development process, the construction and testing of different network designs should be easier and might improve the state of the art in human activity representation.

\section{Conclusions}
\label{sec:conclusions}

The paper presents \MON, a smart home framework aimed at recognising and classifying human activities based on a network of ontologies. 
Each ontology classifies sensory data based on \emph{a priori} defined, context-based knowledge.
In the network, ontologies are pairwise interfaced via general-purpose computational procedures, which are scheduled when specified events occur within ontologies. 
The idea of \MON is to concurrently manage knowledge based on several contexts, which are used to support heterogeneous computational procedures to evaluate activity models. 
\MON adopts the OWL formalism to enforce intelligibility in support of the domain experts who must design and validate activity models.

With \MON, we developed an ontology network for a use case involving ADL.
In this network, an ontology is used to spatially relate sensory data and to trigger (based on events) temporal reasoning on other ontologies, each of which are concerned with a specific activity.
Hence, \MON reasons based on a concurrent hierarchy of contexts, which are represented as small ontologies subjected to DL-based instance checking only when required.
This design allows to
($i$)~decrease computational load, 
($ii$)~assure intelligibility, and 
($iii$)~develop modular and flexible systems that enable an iterative development process involving domain experts.
The network we developed is to be meant at showing the possibilities inherent in the architectural choices allowed by \MON.

A further work is to integrate \MON with the Robot Operating System using~\cite{armor_ws} for accessing OWL ontologies.
Also, we want to exploit~\cite{architecturalSIT} for learning the necessary prior knowledge by interacting with the assisted persons while they are performing relevant activities.
Finally, we aim at implementing an autobiographical memory representing the assisted persons' routine based on the work described in~\cite{buoncompagni2019framework}.

\bibliographystyle{IEEEtran}
\bibliography{IEEEabrv,Human-Activity-Recognition-Models-in-Ontology-Networks}

\newpage
\begin{IEEEbiography}[{\includegraphics[width=1in,height=1.25in,clip,keepaspectratio]{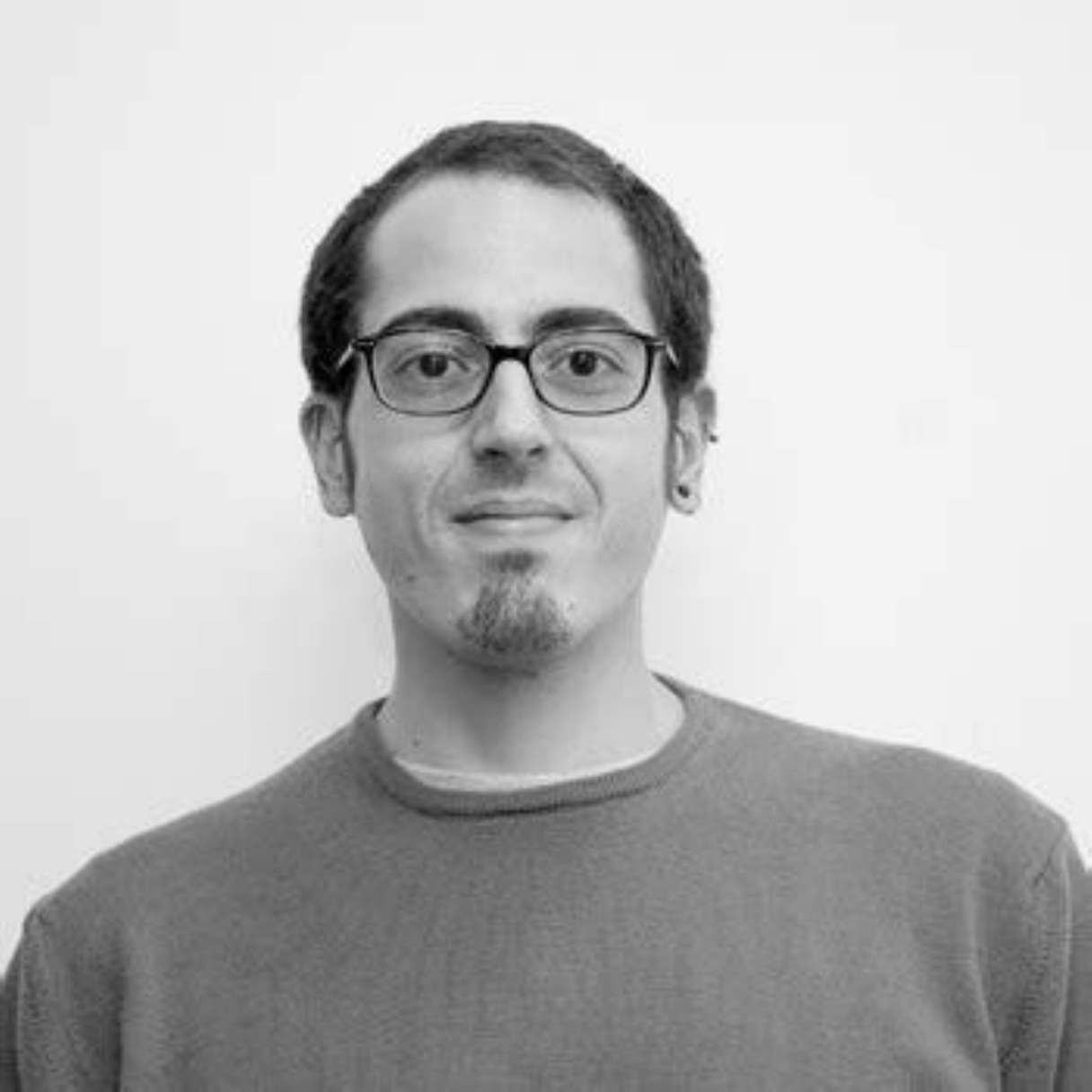}}]{Luca Buoncompagni}
is a researcher in Human-Robot Interaction and Ambient Intelligence, mainly focussed on knowledge representation and architectural perspectives.
His research interests involve human-in-the-loop scenarios and system transparency.
He graduated in Automation and Robotic Engineering in 2014, and he obtained a PhD degree in Robotics and Artificial Intelligence in 2018 with a thesis concerning autobiographic robots memory.
He is currently affiliated with the Department of Informatics, Bioengineering, Robotics, and System Engineering of the University of Genoa (UniGe), and collaborating with Teseo: a spinoff company from UniGe.
\end{IEEEbiography}

\begin{IEEEbiography}[{\includegraphics[width=1in,height=1.25in,clip,keepaspectratio]{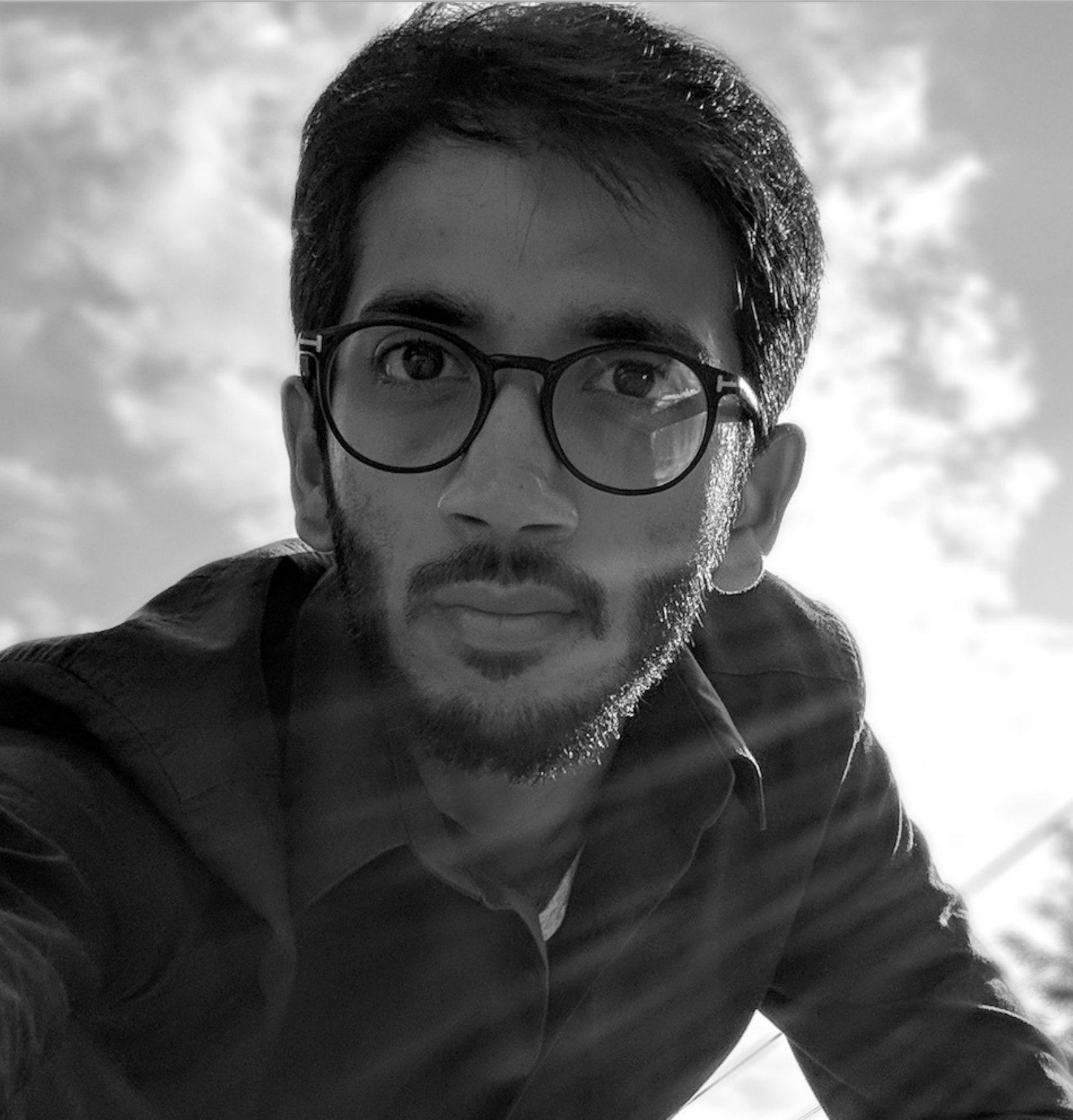}}]{Syed Yusha Kareem}
graduated from EMARO+ (European Master on Advanced RObotics Plus) programme thus receiving dual Master's degrees in robotics engineering - a degree from Centrale Nantes, Nantes, France, in 2016, and a degree from University of Genoa, Genoa, Italy, in 2017. He is currently pursuing the Ph.D. degree in Robotics and Autonomous Systems (RAS) with the Department of Informatics, Bioengineering, Robotics and Systems Engineering (DIBRIS), University of Genoa, Genoa, Italy.
His research interests include ambient intelligence, artificial intelligence based robotics, and human-robot interaction.
\end{IEEEbiography}

\begin{IEEEbiography}[{\includegraphics[width=1in,height=1.25in,clip,keepaspectratio]{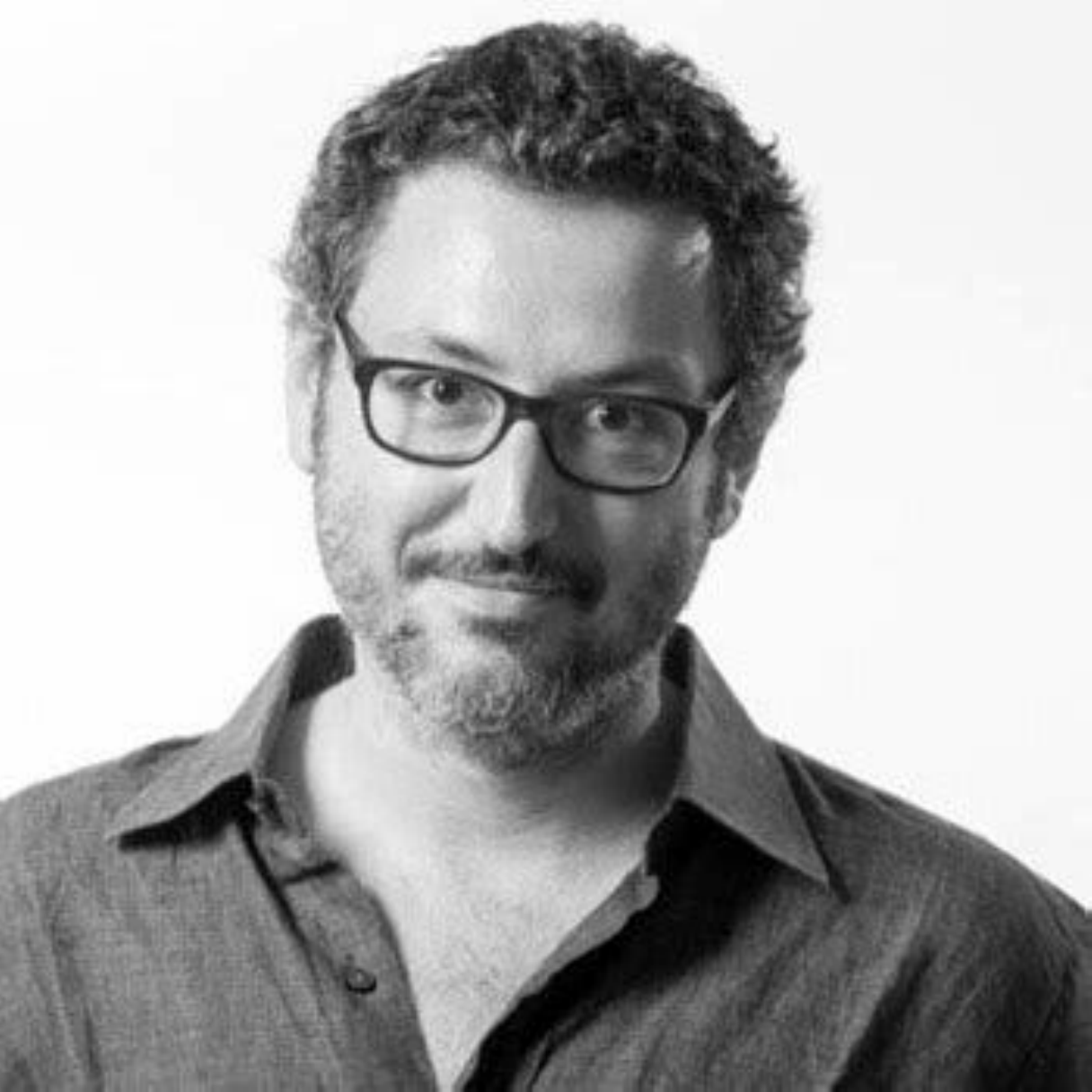}}]{Fulvio Mastrogiovanni}
is an Associate Professor at the University of Genoa (UniGe), Italy, and currently vice-Rector for International Affairs. He got a Laurea Degree and a PhD from UniGe in 2003 and 2008, respectively. Fulvio was a Visiting Professor at institutions in America, Europe, and Asia. Fulvio was part of the Board of Directors of the Italian Association for Artificial Intelligence and is a member of the Board of the PhD School in Bioengineering and Robotics, the Commission for International Relationships, and Program Coordinator of the MSc in Robotics Engineering (part of the Erasmus+ EMARO and JEMARO programs) at UniGe. Fulvio organized international scientific events (RO-MAN, IROS, ERF conference series), and journal special issues. Fulvio teaches courses in robot architectures, human-robot interaction, artificial intelligence for robotics. Fulvio participated in many international funded projects and cooperates with research centers worldwide. He is the founder of Teseo and IOSR, two spinoff companies from UniGe, and an advisor for HIRO Robotics. Fulvio received awards for his scientific and technology transfer activities. Fulvio published more than 170 contributions, including 4 patents.
\end{IEEEbiography}

\vfill
\newpage

\appendices

\begin{figure}[t]
    \centering
    \footnotesize
        \includegraphics[width=.7\textwidth]{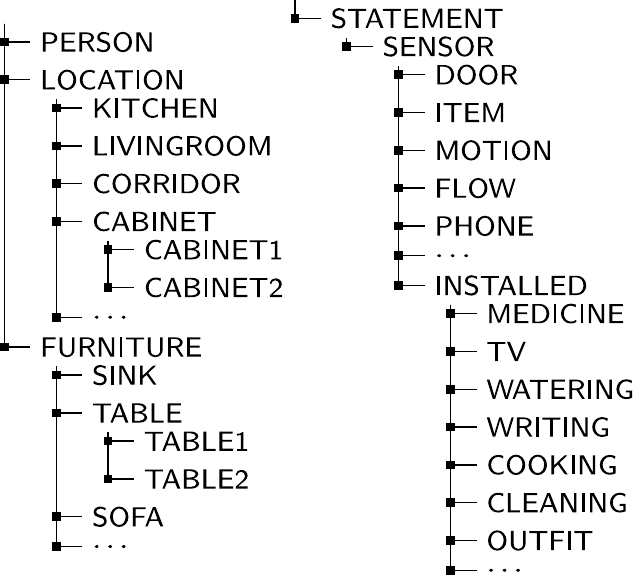}
    \caption{%
        Hierarchies of classes in the ontologies shown as sub-classes, \eg \dltext{SENSOR$\,\sqsubseteq\,$STATEMENT}. 
        It partially represents the prior knowledge that we used to address the scenario presented in Section~\ref{sec:scenario}.
    }
    \label{apx:tab:onto}
\end{figure}

\begin{table}[t]%
    \newcommand{\nl}[1]{\multicolumn{1}{@{}c@{}}{\begin{tabular}[c]{@{}c@{}}#1\end{tabular}}}
    \newcommand{\titem}[4]{%
        \multicolumn{1}{@{}c@{~~~~}}{\begin{tabular}[c]{@{}l@{}}#1\end{tabular}} &
        \multicolumn{1}{@{}l@{~~~~}}{\begin{tabular}[c]{@{}l@{}}#2.\end{tabular}} &
        \multicolumn{1}{@{}l@{~~~~}}{\begin{tabular}[c]{@{}l@{}}#3.\end{tabular}} &
        \multicolumn{1}{@{}l@{}}{\begin{tabular}[c]{@{}l@{}}#4.\end{tabular}} 
    }%
    \newcommand{\titemC}[4]{%
        \multicolumn{1}{@{}c@{~~~~}}{\begin{tabular}[c]{@{}c@{}}#1\end{tabular}} &
        \multicolumn{1}{@{}c@{~~~~}}{\begin{tabular}[c]{@{}c@{}}#2.\end{tabular}} &
        \multicolumn{1}{@{}c@{~~~~}}{\begin{tabular}[c]{@{}c@{}}#3.\end{tabular}} &
        \multicolumn{1}{@{}c@{}}{\begin{tabular}[c]{@{}c@{}}#4.\end{tabular}} 
    }%
    \centering 
    \footnotesize   
    \begin{tabular}{cccc}
        \titemC{Edge\\Name}{Scheduling\\Events}{Required\\Statements}{Provided\\Statements}   \\ \hline
        \titem{$\mathcal{D}$}%
              {~~~At bootstrap}%
              {None}%
              {\dltext{X:SENSOR}\\in $\mathscr{L}$}%
        \\\hline
        \titem{$\mathcal{I}_1$}%
              {~~\dltext{(P,K):isIn}\\[.2em]in $\mathscr{L}$, where\\\dltext{K:KITCHEN},\\\dltext{P:PERSON}}%
              {\dltext{D$_i$:MEDICINE}\\~~~$\sqcap$ \dltext{DOOR},\\\dltext{I$_i$:MEDICINE}\\~~~$\sqcap$ \dltext{ITEM}\\in $\mathscr{L}$}%
              {\emph{the Required}\\\emph{Statements and},\\[.2em]\dltext{N:UPDATE}\\in $\mathscr{T}_1$}%
        \\\hline
        \titem{$\mathcal{I}_2$}%
              {~~\dltext{(P,L):isIn}\\[.2em]in $\mathscr{L}$, where\\\dltext{L:LIVINGROOM}}%
              {\dltext{I$_i$:TV}\\~~~$\sqcap$ \dltext{ITEM},\\in $\mathscr{L}$}%
              {\emph{the Required}\\\emph{Statements and},\\[.2em]\dltext{N:UPDATE}\\in $\mathscr{T}_2$}%
        \\\hline
        \titem{$\mathcal{I}_3$}%
              {~~\dltext{(L,C):isNearTo}\\$\lor$\dltext{(P,S):isNearTo}\\$\lor$\dltext{(P,L):isIn}\\[.2em]in $\mathscr{L}$, where\\\dltext{C:CABINET1},\\\dltext{S:SINK}}%
              {\dltext{D$_i$:WATERING}\\~~~$\sqcap$ \dltext{CABINET1},\\\dltext{M$_i$:WATERING}\\~~~$\sqcap$ \dltext{MOTION},\\\dltext{F$_i$:SINK}\\in $\mathscr{L}$}%
              {\emph{the Required}\\\emph{Statements and},\\[.2em]\dltext{N:UPDATE}\\in $\mathscr{T}_3$}%
        \\\hline
        \titem{$\mathcal{I}_4$}%
              {~~\dltext{(P,T2):isNearTo}\\[.2em]in $\mathscr{L}$, where\\\dltext{T2:TABLE2}}%
              {\dltext{H$_i$:PHONE}\\in $\mathscr{L}$}%
              {\emph{the Required}\\\emph{Statements and},\\[.2em]\dltext{N:UPDATE}\\in $\mathscr{T}_4$}%
        \\\hline
        \titem{$\mathcal{I}_5$}%
              {~~\dltext{(P,T1):isNearTo}\\[.2em]in $\mathscr{L}$, where\\\dltext{T1:TABLE1}}%
              {\dltext{I$_i$:WRITING}\\~~~$\sqcap$ \dltext{ITEM}\\in $\mathscr{L}$}%
              {\emph{the Required}\\\emph{Statements and},\\[.2em]\dltext{N:UPDATE}\\in $\mathscr{T}_5$}%
        \\\hline
        \titem{$\mathcal{I}_6$}%
              {~~\dltext{(P,K):isIn}\\[.2em]in $\mathscr{L}$}%
              {\dltext{D$_i$:COOKING}\\~~~$\sqcap$ \dltext{DOOR},\\\dltext{I$_i$:COOKING}\\~~~$\sqcap$ \dltext{ITEM}\\in $\mathscr{L}$}%
              {\emph{the Required}\\\emph{Statements and},\\[.2em]\dltext{N:UPDATE}\\in $\mathscr{T}_6$}%
        \\\hline
        \titem{$\mathcal{I}_7$}%
              {~~\dltext{(P,L):isIn}\\$\lor$\dltext{(P,K):isIn}\\[.2em]in $\mathscr{L}$}%
              {\dltext{D$_i$:CLEANING}\\~~~$\sqcap$ \dltext{DOOR},\\\dltext{M$_i$:CLAENING}\\~~~$\sqcap$ \dltext{MOTION}\\in $\mathscr{L}$}%
              {\emph{the Required}\\\emph{Statements and},\\[.2em]\dltext{N:UPDATE}\\in $\mathscr{T}_7$}%
        \\\hline
        \titem{$\mathcal{I}_8$}%
              {~~\dltext{(P,R):isIn}\\$\lor$\dltext{(P,T1):isNearTo}\\$\lor$\dltext{(P,O):isNearTo}\\[.2em]in $\mathscr{L}$, where\\\dltext{R:CORRIDOR},\\\dltext{O:SOFA}}%
              {\dltext{D$_i$:OUTFIT}\\~~~$\sqcap$ \dltext{DOOR},\\\dltext{M$_i$:OUTFIT}\\~~~$\sqcap$ \dltext{MOTION}\\in $\mathscr{L}$}%
              {\emph{the Required}\\\emph{Statements and},\\[.2em]\dltext{N:UPDATE}\\in $\mathscr{T}_8$}%
        \\\hline
        \titem{\nl{$\mathcal{M}_a$\\$a{\in}[1,8]$}}%
              {~~\dltext{N:UPDATED}\\in $\mathscr{T}_a$}%
              {\dltext{X:SENSOR}\\in $\mathscr{T}_a$}%
              {\vs{A_a} in $\mathscr{T}_a$}%
    \end{tabular}   
    \caption{The design of the interface among computational procedures and ontologies for the network shown in Figure~\ref{fig:setup}.}
    \label{apx:tab:proc}
\end{table}

\section{An Ontology Network Configuration}
\label{apx:network}

This section details the implementation of the network depicted in Figure~\ref{fig:setup}, and developed based on a specific scenario as presented in Section~\ref{sec:experiment}.

\subsection{Ontologies}
\label{sec:apx:onto}
Figure~\ref{apx:tab:onto} shows relevant prior knowledge encoded inside the ontologies of the network, and it shows two disjoint sets of classes and sub-classes.
The left-hand side concerns the localisation of a person in a room and their proximity to furniture.
Instead, the right-hand side represents a hierarchy of classes concerning sensors and the type of activity they are supposed to monitor in a particular installation.
If not explicitly stated, sub-classes are not disjoint, \eg we allow the presence sensor \dltext{D$_7$:DOOR} to be an instance of the \dltext{MEDICINE} and \dltext{COOKING} classes at the same time.

Figure~\ref{fig:setup} shows two types of ontologies, \ie the spatial ontology $\mathscr{L}$ and the activity ontology $\mathscr{T}_a$, which is instantiated for each activity $a\in[1,8]$.
In the $\mathscr{L}$ ontology, we encode both hierarchies of classes shown in Figure~\ref{apx:tab:onto}, while each $\mathscr{T}_a$ shares with $\mathscr{L}$ only a part of the hierarchy on the right-hand side.
In particular, $\mathscr{T}_a$ encodes only the part related to the statements provided by $\mathcal{I}_a$, as shown in Table~\ref{apx:tab:proc} (and detailed further in the next Section).
Remarkably, each ontology $\mathscr{T}_a$ also contains other prior knowledge, \eg an \dltext{UPDATE} concept to classify events for synchronising computational procedures, or a dedicated representation of \dltext{TIME} based on Allen's Algebra~\cite{allen1983}.

In $\mathscr{L}$, we spatially contextualise knowledge by defining 
\begin{align}
    \tdl{PERSON}    \doteq&~ \tdl{isIn.LOCATION}\,{\sqcap}\,\tdl{isNearTo.FURNITURE}, \notag\\
    \tdl{STATEMENT} \doteq&~ \tdl{hasState.}\mathbb{B} \sqcap \tdl{hasTime.}\mathbb{N}^{+}, \notag\\
    \tdl{SENSOR}    \doteq&~ \tdl{STATEMENT} \sqcap {\geqslant}1\,\tdl{isIn.LOCATION} \notag\\
                          &~ \sqcap~{\geqslant}0\,\tdl{isNearTo.FURNITURE}, \notag\\
    \tdl{LOCATION}  \neg  &  \tdl{FURNITURE},\;\; 
    \tdl{SENSOR}    \neg     \tdl{PERSON}.    
\end{align}
Here, $\mathbb{B}$ represents the Boolean \emph{concrete} concept, while $\mathbb{N}^{+}$ represents integer positive numbers in accordance with Section~\ref{sec:statement}.
Based on this representation, we consider an instance \dltext{P:PERSON} having properties \dltext{(P,X):isIn} and \dltext{(P,Y):isNearTo} and an instance \dltext{S:SENSOR}, which can have a high state, \ie \dltext{(S,$\top$):hasState}.
In other words, given sensors with a $\top$ state, we spatially localise a person based on the properties \dltext{(S,X):isIn} and \dltext{(S,Y):isNearTo}.

The \dltext{INSTALLED} class describes sets of sensors that are related to a particular activity, and it is used to interface computational procedures as shown in Table~\ref{apx:tab:proc}.
The sets of sensors are the following,
\begin{align}
    \label{apx:eq:activitySensor}
    \left\{\tdl{D$_7$, I$_4$, I$_6$, I$_7$}\right\}                                               & \tdl{:MEDICINE}, \notag\\
    \left\{\tdl{I$_5$, I$_3$}\right\}                                                             & \tdl{:TV},       \notag\\
    \left\{\tdl{D$_{11}$, F$_2$, F$_3$, M$_{6}$, M$_{7}$, $\ldots$, M$_{14}$}\right\}             & \tdl{:WATERING}, \notag\\
    \left\{\tdl{I$_8$, I$_9$}\right\}                                                             & \tdl{:WRITING},  \notag\\
    \left\{\tdl{D$_8$, D$_9$, D$_{10}$, I$_1$, I$_2$, I$_7$}\right\}                              & \tdl{:COOKING},  \notag\\
    \left\{\tdl{D$_{11}$, M$_6$, M$_7$, $\ldots$, M$_{10}$, M$_{16}$, M$_{17}$, M$_{18}$}\right\} & \tdl{:CLEANING}, \notag\\
    \left\{\tdl{D$_{12}$, M$_3$, M$_4$, $\ldots$, M$_9$, M$_{21}$, M$_{22}$, M$_{23}$}\right\}    & \tdl{:OUTFIT}.   
\end{align}
We denote \dltext{SENSOR} instances as \dltext{D$_i$:DOOR}, \dltext{I$_i$:ITEM}, \dltext{M$_i$:MOTION}, \dltext{F$_i$:FLOW} and \dltext{H:PHONE}, as shown in Figure~\ref{fig:home}. 
The activity of answering the phone is not listed in \eqref{apx:eq:activitySensor} because it is based on a single sensor, \ie \dltext{H}.

\subsection{Computational Procedures}
\label{sec:apx:proc}
The computational procedures are interfaced with the ontologies as shown in Table~\ref{apx:tab:proc}, \ie based on a scheduling event and given some required statements the computation of procedures is triggered, in accordance with Section~\ref{sec:overview}.

Figure~\ref{fig:setup} shows three types of procedures.
$\mathcal{D}$ creates instances of the \dltext{SENSOR} class in the $\mathscr{L}$ ontology, therefore simulating a stream of data.
$\mathcal{D}$ updates the reasoner of the spatial ontology $\mathscr{L}$ each time it creates a new sensor statement in the ontology.
Since $\mathscr{L}$ represents a context that does not involve time, old statements generated by a sensor are overwritten when the sensors provide new statements.

$\mathcal{I}_a$ reacts to events related to the position of the person in accordance with  Table~\ref{apx:tab:proc}.
When $\mathcal{I}_a$ is scheduled, it moves statements that are relevant for the $a$-th activity from $\mathscr{L}$ to $\mathscr{T}_a$.
The statements moved by $\mathcal{I}_a$ are retrieved by means of a sub-class of the \dltext{INSTALLED} concept that is relevant for a specific activity, as shown in \eqref{apx:eq:activitySensor} and Figure~\ref{apx:tab:onto}.
$\mathcal{I}_a$ also creates a statement \dltext{N$^\top$} (in $\mathscr{T}_a$) with the purpose to generate an event that would trigger the computation of $\mathcal{M}_a$.
In order to generate this event we encode in $\mathscr{T}_a$ prior knowledge such that \dltext{N:UPDATE} if $\bm{s}(\tdl{N})=\top$, while if it is $\bot$, \dltext{N} would not be an instance of the \dltext{UPDATE} class.

$\mathcal{M}_a$ is scheduled when in $\mathscr{T}_a$ it does exist a statement \dltext{N} classified as \dltext{UPDATE}.
When $\mathcal{M}_a$ starts, it resets \dltext{N}, \ie it sets $\bm{s}(\tdl{N})=\bot$ in $\mathscr{T}_a$, and \dltext{N} is not an instance of the \dltext{UPDATE} class anymore.
Then, $\mathcal{M}_a$ computes the $a$-th activity model based on \dltext{SENSOR} statements in $\mathscr{T}_a$, which are collected over time.
If the model recognises the activity, the aggregated statement \vs{A_a} will be stored in $\mathscr{T}_a$, and all the old statements are removed from that ontology.

Since the models to recognise activities are based on SWRL rules (which are discussed in Appendix~\ref{apx:models}), $\mathcal{M}_a$ can update the reasoner of $\mathscr{T}_a$ to evaluate whether an activity is performed.
However, SWRL rules do not support certain operations required to implement the statement's algebra, as discussed in Section~\ref{sec:statement}.
Therefore, we design $\mathcal{M}_a$ to perform some computation before evaluating the SWRL rule.

\section{Activity Models}
\label{apx:models}

This Appendix details the implementation of activity models shown in Figure~\ref{fig:models}.
The models are specifically designed for the scenario presented in Section~\ref{sec:experiment}, depicted in Figure~\ref{fig:setup} and detailed in Appendix~\ref{apx:network}.

Section~\ref{apx:sec:swrlPremier} introduces SWRL rules, and it discusses some of their aspects related to \MON.
From Section~\ref{apx:sec:a1} to Section~\ref{apx:sec:a8}, \ie for each activity model shown in Figure~\ref{fig:models}, we provide 
($i$)~a formalisation with statement's algebra, 
($ii$)~an explanation with respect to prior knowledge, and
($iii$)~an implementation based on logic and rules.

\subsection{SWRL Rules and \MON Statements}
\label{apx:sec:swrlPremier}

A SWRL rule is an expression made by a conjunction of logic atoms, and if the expression is satisfied, then an implication is deduced.
\begin{equation}
    a \wedge b \wedge c \implies d.
\end{equation}
Therefore, if all the atoms $\{a, b, c\}$ are satisfied in an ontology, $d$ is deduced, \ie the atom $d$ will exist in the ontology.

SWRL atoms can represent the classification of an instance \dltext{X} in a concept of the ontology, \eg the DL expression \dltext{X:KITCHEN} is addressed as \dltext{KITCHEN(X)}.
Furthermore, an atom may be concerned with the properties beyween two instances \dltext{X,Y}, \eg the DL expression \dltext{(X,Y):isIn} is identified as \dltext{isIn(X,Y)}.

Atoms can be also used to perform basic algebraic computations and comparisons, and they support the definition of variables, which are indicated with the `\dltext{?}' symbol.
To simplify the notation, we replace the atoms \dltext{greaterThan(?g,k)} with $(\tdl{?g}\leqslant k)$, and \dltext{sum(?r,k,?d)} with $(\tdl{?r}\gets k+\tdl{?d})$, where the arrow identifies an assignment, and $k\in\mathbb{N}^+$ might be a constant value.

Those types of atoms are evaluated by some OWL reasoner (\eg Pellet) to compute fluent models.
In particular, SWRL rules are computed by checking all symbols' permutations that are consistent within the classes and properties encoded in an ontology.

SWRL rules are applied to activity ontologies $\mathscr{T}_a$ that are not concerned with all the sensory data in the considered scenario.
Instead, ontologies are concerned with only those sensor statements that are selected by $\mathcal{I}_a$ based on spatial contextualisation (Table~\ref{apx:tab:proc}).
For clarity, in the models formalised with the statement's algebra, we explicitly indicate the sensor with respect to Figure~\ref{fig:home}, but this is not true in the notation of the SWRL rule.
This is because $\mathscr{T}_a$ represents a temporal context and not a spatial one, therefore, it does not encode the position of the sensors as it is done in $\mathscr{L}$.
In other words, we assume that $\mathscr{T}_a$ encodes only sensor statements that are relevant for the $a$-th activity model.
Consequently, there is the need to contextualise knowledge only in terms of sensors type and time intervals.

\begin{table}
    \centering
    \footnotesize
    \begin{tabular}{c@{~~}|@{~~}c@{~~~~}c@{~~~~}c@{~~~~}c@{~~~~}c@{~~~~}c@{~~~~}c@{~~~~}c}
                   & $\alpha_1$ & $\alpha_2$ & $\alpha_3$ & $\alpha_4$ & $\alpha_5$ & $\alpha_6$ & $\alpha_7$ & $\alpha_8$ \\\hline
        $\alpha_1$ & 0.95       & 0          & 0          & 0          & 0          & 0          & 0          & 0          \\
        $\alpha_2$ & 0          & 0.95       & 0.05       & 0          & 0          & 0          & 0.05       & 0          \\
        $\alpha_3$ & 0          & 0          & 0.7        & 0          & 0          & 0          & 0.05       & 0          \\ 
        $\alpha_4$ & 0          & 0          & 0          & 1          & 0          & 0          & 0.05       & 0          \\
        $\alpha_5$ & 0          & 0          & 0          & 0          & 0.8        & 0          & 0          & 0          \\
        $\alpha_6$ & 0          & 0          & 0          & 0          & 0          & 0.7        & 0.05       & 0          \\
        $\alpha_7$ & 0          & 0          & 0.25       & 0          & 0          & 0          & 0.8        & 0          \\
        $\alpha_8$ & 0          & 0          & 0          & 0          & 0          & 0          & 0          & 0.95       \\
     unclassified  & 0.05       & 0.05       & 0          & 0          & 0.2        & 0.3        & 0          & 0.05 
    \end{tabular}
    \caption{Confusion matrix of the activity recognition rate obtained with the ontology network designed in Section~\ref{sec:experiment}.}
    \label{fig:confusion}
\end{table}

Table~\ref{fig:confusion} shows the recognition performance of the following models.
From this Table, the Table~\ref{tab:Fmeasure} is deduced.

\subsection{Activity Model \texorpdfstring{$\alpha_1$}{\textalpha\textoneinferior}}
\label{apx:sec:a1}

The first activity concerns filling the medical dispenser.
An implementation based on statement's algebra, and applied to the ontology $\mathscr{T}_1$, is%
\footnote{In \eqref{eq:model_a1T} we had considered a simplified case wherebe statements related to the object $I^\bot_7$ do not exist in the $\mathscr{T}_1$ context.}
\begin{align}
    \label{eq:apx:a1}
    \vs{A}_1 \vDash&~ \big(\big(\vs{T}\land\top\big) + \delta_1\big) \leqslant \big(\vs{R}\land\top\big),
    \;\;\text{where}\notag\\
  	\vs{T} &\vDash D^\top_7 \leqslant \big(\big(I^\bot_4 \land I^\bot_6\big) \lor  \big(I^\bot_4 \land I^\bot_7\big) \lor \big(I^\bot_6 \land I^\bot_7\big)\big),\notag\\
	\vs{R} &\vDash \big(\big(I^\top_4 \land I^\top_6\big) \lor  \big(I^\top_4 \land I^\top_7\big) \lor \big(I^\top_6 \land I^\top_7\big)\big) \leqslant D^\bot_7.
\end{align}

\noindent
The activity $\alpha_1$ is recognised based on a statement \vs{A}$_1$ that can be expressed as 
\expl{two objects were taken (\vs{T}), \\then after $\delta_1$ time units, they were released (\vs{R}).}
Here, \vs{T} and \vs{R} can be expressed respectively as
\expl{the door was opened, then the objects were absent.'',\\
    ``the objects were present, then the door was closed.}
The two \emph{objects} are associated on the basis of prior knowledge, \eqref{apx:eq:activitySensor} and Table~\ref{apx:tab:proc}, \ie as
\expl{medicinal items are in a cabinet of the kitchen.}

\noindent
The SWRL-based implementation of \eqref{eq:apx:a1} is the following rule 
\begin{align}
    &\tdl{DOOR(?O)} \land \tdl{hasState(?O,$\top$)} \land \tdl{hasTime(?O, ?t$_O$)}                               ~\land\notag\\
    &\tdl{ITEM(?I)} \land \tdl{hasState(?I,$\bot$)} \land \tdl{hasTime(?I, ?t$_I$)}                               ~\land\notag\\    
    &\tdl{ITEM(?J)} \land \tdl{hasState(?J,$\bot$)} \land \tdl{hasTime(?J, ?t$_J$)}                               ~\land\notag\\
    &(\tdl{I}\neq\tdl{J}) \land (\tdl{?t$_O$} \leqslant \tdl{?t$_I$}) \land (\tdl{?t$_O$} \leqslant \tdl{?t$_J$}) ~\land\notag\\[.3em]
    &\tdl{DOOR(?C)} \land \tdl{hasState(?C,$\bot$)} \land \tdl{hasTime(?C, ?t$_C$)}                               ~\land\notag\\
    &\tdl{ITEM(?W)} \land \tdl{hasState(?W,$\top$)} \land \tdl{hasTime(?W, ?t$_W$)}                               ~\land\notag\\    
    &\tdl{ITEM(?Q)} \land \tdl{hasState(?Q,$\top$)} \land \tdl{hasTime(?Q, ?t$_Q$)}                               ~\land\notag\\
    &(\tdl{W}\neq\tdl{Q}) \land (\tdl{?t$_W$} \leqslant \tdl{?t$_C$}) \land (\tdl{?t$_Q$} \leqslant \tdl{?t$_O$}) ~\land\notag\\[.3em]
    &(\tdl{?t} \gets \tdl{?t$_O$} + \delta_1) \land (\tdl{?t} \leqslant \tdl{?t$_C$})                                   \notag\\
    &~~~~\implies \tdl{(\vs{A_1},$\top$):hasState} \land \tdl{(\vs{A_1},?t$_C$):hasTime}. 
\end{align}

\subsection{Activity Model \texorpdfstring{$\alpha_2$}{\textalpha\texttwoinferior}}
\label{apx:sec:a2}

The second activity is watching a DvD, and an implementation based on statement's algebra involving $\mathscr{T}_2$ is shown in \eqref{eq:mm2}.
The statement \vs{A}$_2$, which represents the recognition of the $\alpha_2$ activity, can be expressed as 
\expl{an object was taken, then after $\delta_2$ time units,\\it was released,}
where the \emph{object} is associated with prior knowledge concerning \dltext{TV}, \dltext{ITEM}, \dltext{LIVINGROOM} and related furnitures, \eg \dltext{SOFA}, in accordance with \eqref{apx:eq:activitySensor} and Table~\ref{apx:tab:proc}.

The implementation of \eqref{eq:mm2} based on SWRL is a rule
\begin{align}
    &\tdl{ITEM(?T)} \land \tdl{hasState(?T,$\bot$)} \land \tdl{hasTime(?T, ?t$_T$)}   ~\land\notag\\    
    &\tdl{ITEM(?R)} \land \tdl{hasState(?R,$\top$)} \land \tdl{hasTime(?R, ?t$_R$)}   ~\land\notag\\
    &(\tdl{?t} \gets \tdl{?t$_T$} + \delta_2) \land (\tdl{?t} \leqslant \tdl{?t$_R$})  ~\notag\\
    &~~~~\implies \tdl{(\vs{A_2},$\top$):hasState} \land \tdl{(\vs{A_2},?t$_R$):hasTime}. 
\end{align}

\subsection{Activity Model \texorpdfstring{$\alpha_3$}{\textalpha\textthreeinferior}}
\label{apx:sec:a3}

The third activity concerns watering plants, it involves $\mathscr{T}_3$, and \eqref{eq:mm3} shows its definition with the statement's algebra.
The statement related to this model (\ie \vs{A}$_3$) can be expressed as 
\expl{the door was opened and the sink was used, then the person stayed near a plant for $\delta_3$ time units, and near another plant for $\delta_4$ time units, then the door got closed.}
Here the furniture involved is represented with respect to prior knowledge representing a \dltext{CABINET} door, the \dltext{SINK} and two areas in the \dltext{LIVINGROOM} where the two plants are supposed to be.
All those sensor statements are classified in $\mathscr{L}$ through the class \dltext{WATERING} shown in \eqref{apx:eq:activitySensor} and Table~\ref{apx:tab:proc}.
With respect to Figure~\ref{apx:tab:onto}, we consider additional prior knowledge
\begin{align}
           \{\tdl{M$_6$},\ldots,\tdl{M$_9$}\}&\tdl{:PLANT1}\sqsubseteq(\tdl{WATERING}\sqcap\tdl{MOTION}), \notag\\
      \{\tdl{M$_{10}$}\ldots,\tdl{M$_{14}$}\}&\tdl{:PLANT2}\sqsubseteq(\tdl{WATERING}\sqcap\tdl{MOTION}).
\end{align}

\noindent
The SWRL-based implementation of \eqref{eq:mm3} is
\begin{align}
    \label{apx:eq:swrl3}
    &\tdl{DOOR(?O)} \land \tdl{hasState(?O,$\top$)} \land \tdl{hasTime(?O, ?t$_O$)}                             ~\land\notag\\    
    &\tdl{DOOR(?C)} \land \tdl{hasState(?C,$\bot$)} \land \tdl{hasTime(?C, ?t$_C$)}                             ~\land\notag\\
    &\tdl{FLOW(?F)} \land \tdl{hasState(?F,$\top$)} \land \tdl{hasTime(?F, ?t$_F$)}                             ~\land\notag\\
    &(\tdl{?t$_{O}$} \leqslant \tdl{?t$_C$}) \land (\tdl{?t$_{O}$} \leqslant \tdl{?t$_F$})                      ~\land\notag\\[.3em]
     &\tdl{WATERED(?\vs{G})} \land \tdl{hasState(?\vs{G},$\top$)} \land \tdl{hasTime(?\vs{G}, ?t$_{\vs{G}}$)} ~\land\notag\\
     &\tdl{WATERED(?\vs{E})} \land \tdl{hasState(?\vs{E},$\top$)} \land \tdl{hasTime(?\vs{E}, ?t$_{\vs{E}}$)} ~\land\notag\\   
     &(\vs{G}\neq\vs{E}) \land (\tdl{?t$_{F}$} \leqslant \tdl{?t$_{\vs{G}}$})                                  \land 
      (\tdl{?t$_F$} \leqslant \tdl{?t$_{\vs{E}}$})                                                            ~\land\notag\\
     &(\tdl{?t$_{\vs{G}}$} \leqslant \tdl{?t$_C$}) \land (\tdl{?t$_{\vs{E}}$} \leqslant \tdl{?t$_C$})               \notag\\[.5em]    
    &~~~~\implies \tdl{(\vs{A_3},$\top$):hasState} \land \tdl{(\vs{A_3},t$_C$):hasTime}. 
\end{align}

\noindent
Since SWRL respects the open-world assumption, we used further computation to perform the statements convolutions \eqref{eq:convolution}.
This computation is performed by $\mathcal{M}_3$ before applying the above rule.
In particular, $\mathcal{M}_3$ queries all the timestamps associated with the statements \dltext{$M_i$:PLANT1$\sqsubseteq$SENSOR} in $\mathscr{T}_3$, which contain the visits to locations of the plants; as imported by $\mathcal{I}_a$ in accordance with Appendix~\ref{apx:network}.
Then, $\mathcal{M}_3$ computes the minimum ($t^-$) and maximum ($t^+$) values of the queried statements, and it counts their number ($n$).
If $t^- + \delta_3 \leqslant t^+$ and $n \leqslant h_3$, then $\mathcal{M}_3$ generates a statement $\vs{G}$ with a $\top$ state and a timestamp equal to $t^+$.
With an equivalent approach based on \dltext{$M_i$:PLANT2} but with different parameters $\delta_4$ and $h_4$, $\mathcal{M}_3$ also generates a statement \vs{E}.
$\mathcal{M}_3$ generates \vs{G} and \vs{E} in $\mathscr{T}_3$ as instances of the \dltext{WATERED$\sqsubseteq$WATERING$\sqsubseteq$SENSOR} class, which represents that some water has been given to a plant.
Then, $\mathcal{M}_3$ computes the activity model by updating the reasoner of $\mathscr{T}_3$, which evaluates \eqref{apx:eq:swrl3}.

\subsection{Activity Model \texorpdfstring{$\alpha_4$}{\textalpha\textfourinferior}}

The fourth activity is answering to the Phone, which we define with statement's algebra in $\mathscr{T}_4$ as
\begin{align}
    \label{eq:apx:a4}
    \vs{A}_4 &\vDash \big(H^\top_1 + \delta_5\big) \leqslant H^\bot_1.
\end{align}
\vs{A}$_4$ can be expressed as in Appendix~\ref{apx:sec:a2}, where the object is associated with prior knowledge concerning \dltext{H:PHONE}.
The SWRL-based implementation of \eqref{eq:apx:a4} is the rule
\begin{align}
    &\tdl{PHONE(?T)} \land \tdl{hasState(?T,$\top$)} \land \tdl{hasTime(?T, ?t$_T$)}   ~\land\notag\\    
    &\tdl{PHONE(?R)} \land \tdl{hasState(?R,$\bot$)} \land \tdl{hasTime(?R, ?t$_R$)}   ~\land\notag\\
    &(\tdl{?t} \gets \tdl{?t$_T$} + \delta_5) \land (\tdl{?t} \leqslant \tdl{?t$_R$})  ~\notag\\
    &~~~~\implies \tdl{(\vs{A_4},$\top$):hasState} \land \tdl{(\vs{A_4},t$_R$):hasTime}. 
\end{align}

\subsection{Activity Model \texorpdfstring{$\alpha_5$}{\textalpha\textfiveinferior}}

The fifth activity concerns writing a card, it involves $\mathscr{T}_5$, and \eqref{eq:mm5} shows the related model based in statement's algebra,
which outcome (\ie \vs{A}$_5$) can be expressed as
\expl{two objects were taken, \\then after $\delta_6$ and $\delta_7$ time units, they were released.}
Here, the two \emph{objects} are associated with prior knowledge that spatially describes \dltext{WRITING} in $\mathscr{L}$, as shown in \eqref{apx:eq:activitySensor} and Table~\ref{apx:tab:proc}.
The implementation of \eqref{eq:mm5} based on SWRL is the rule
\begin{align}
    &\tdl{ITEM(?I)} \land \tdl{hasState(?I,$\bot$)} \land \tdl{hasTime(?I, ?t$_I$)}   ~\land\notag\\    
    &\tdl{ITEM(?J)} \land \tdl{hasState(?J,$\top$)} \land \tdl{hasTime(?J, ?t$_J$)}   ~\land\notag\\
    &(\tdl{?t} \gets \tdl{?t$_I$} + \delta_6) \land (\tdl{?t} \leqslant \tdl{?t$_J$}) ~\land\notag\\[.3em]
    &\tdl{ITEM(?W)} \land \tdl{hasState(?W,$\bot$)} \land \tdl{hasTime(?W, ?t$_W$)}               ~\land\notag\\    
    &\tdl{ITEM(?Q)} \land \tdl{hasState(?Q,$\top$)} \land \tdl{hasTime(?Q, ?t$_Q$)}               ~\land\notag\\
    &(\tdl{?f} \gets \tdl{?t$_W$} + \delta_7) \land (\tdl{?f} \leqslant \tdl{?t$_Q$})             ~\land\notag\\[.3em]
    &(\tdl{?t$_J$} \leqslant \tdl{?t$_Q$}) \land (\tdl{I}\neq\tdl{W}) \land (\tdl{J}\neq\tdl{Q})    \notag\\
    &~~~~\implies \tdl{(\vs{A_5},$\top$):hasState} \land \tdl{(\vs{A_5},t$_Q$):hasTime}.
\end{align}

\subsection{Activity Model \texorpdfstring{$\alpha_6$}{\textalpha\textsixinferior}}
\label{apx:sec:a6}

The sixth activity concerns the preparation of a meal, and it is defined with statement's algebra in $\mathscr{T}_6$ as
\begin{align} 
    \label{eq:mm6}
    \vs{A}_6 \vDash&~ \big(D^\top_8 \lor D^\top_9 \lor D^\top_{10}\big) \leqslant \big(\vs{C} \land \vs{C}\big) \leqslant \big(D^\top_8 \lor D^\top_9 \lor D^\top_{10}\big),\notag\\
    \text{where}\;
    \vs{C}  &\vDash  \big(\big(I^\bot_1 \lor I^\bot_2 \lor I^\bot_7\big) + \delta_8\big) \leqslant \big(I^\top_1 \lor I^\top_2 \lor I^\top_7\big).
\end{align}

\noindent
\vs{A}$_6$ can be expressed as 
\expl{doors were opened, two objects were taken,\\then after $\delta_8$ time units the objects were released,\\then the doors got closed,}
where the two \emph{objects} are associated to prior knowledge concerning sensors related to \dltext{COOKING} and their locations, \ie \eqref{apx:eq:activitySensor} and Table~\ref{apx:tab:proc}.
   
The implementation of \eqref{eq:mm6} based on SWRL rules is the following 
\begin{align}
    &\tdl{DOOR(?O)} \land \tdl{hasState(?O,$\top$)} \land \tdl{hasTime(?O, ?t$_O$)}    ~\land\notag\\    
    &\tdl{DOOR(?C)} \land \tdl{hasState(?C,$\bot$)} \land \tdl{hasTime(?C, ?t$_C$)}    ~\land\notag\\
    &(\tdl{?t$_O$} \leqslant \tdl{?t$_C$})                                             ~\land\notag\\[.3em]
    &\tdl{ITEM(?I)} \land \tdl{hasState(?I,$\bot$)} \land \tdl{hasTime(?I, ?t$_I$)}    ~\land\notag\\    
    &\tdl{ITEM(?J)} \land \tdl{hasState(?J,$\top$)} \land \tdl{hasTime(?J, ?t$_J$)}    ~\land\notag\\
    &(\tdl{?t} \gets \tdl{?t$_I$} + \delta_8) \land (\tdl{?t} \leqslant \tdl{?t$_J$})  ~\land\notag\\[.3em]
    &\tdl{ITEM(?W)} \land \tdl{hasState(?W,$\bot$)} \land \tdl{hasTime(?W, ?t$_W$)}    ~\land\notag\\    
    &\tdl{ITEM(?Q)} \land \tdl{hasState(?Q,$\top$)} \land \tdl{hasTime(?Q, ?t$_Q$)}    ~\land\notag\\
    &(\tdl{?f} \gets \tdl{?t$_W$} + \delta_7) \land (\tdl{?f} \leqslant \tdl{?t$_Q$})  ~\land\notag\\[.3em]
    &(\tdl{I}\neq\tdl{W}) \land (\tdl{J}\neq\tdl{Q})                                   ~\land
     (\tdl{?t$_O$} \leqslant \tdl{?t$_I$}) \land (\tdl{?t$_O$} \leqslant \tdl{?t$_W$}) ~\land\notag\\ 
    &(\tdl{?t$_J$} \leqslant \tdl{?t$_C$}) \land (\tdl{?t$_Q$} \leqslant \tdl{?t$_C$})       \notag\\
    &~~~~\implies \tdl{(\vs{A_6},$\top$):hasState} \land \tdl{(\vs{A_6},t$_R$):hasTime}. 
\end{align}

\subsection{Activity Model \texorpdfstring{$\alpha_7$}{\textalpha\textseveninferior}}
\label{apx:sec:a7}

The seventh activity is cleaning the apartment, which is defined using statement's algebra in $\mathscr{T}_7$, as
\begin{align}
    \label{eq:apx:a7}
    \vs{A}_7 \vDash&~ D^\top_{11} \leqslant \big(\vs{H} \land \vs{Q}\big) \leqslant D^\bot_{11},
        \;\;\text{where}\notag\\
    \vs{H}  &\vDash   M_{\{6:10\}}\mcirc{h_9}\delta_9,\notag\\
    \vs{H}  &\vDash   M_{\{16:18\}}\mcirc{h_{10}}\delta_{10},
\end{align}

\noindent
\vs{A}$_7$ can be expressed as 
\expl{the door was opened, then the person stayed $\delta_9$ time units in the leaving room and $\delta_{10}$ time units in the kitchen,\\then the door got closed.}
Here, the door is associated with prior knowledge concerning a \dltext{CABINET}, while $M_{\{6:10\}}$ are related to the \dltext{LIVINGROOM}, and $M_{\{16:18\}}$ to the \dltext{KITCHEN}.
All those sensors are represented by the class \dltext{CLEANING} in $\mathscr{L}$, in accordance with \eqref{apx:eq:activitySensor} and Table~\ref{apx:tab:proc}.
The SWRL-based implementation of \eqref{eq:apx:a7} applied to $\mathscr{T}_7$ is
\begin{align}
    &\tdl{DOOR(?O)} \land \tdl{hasState(?O,$\top$)} \land \tdl{hasTime(?O, ?t$_O$)}                              ~\land\notag\\    
    &\tdl{DOOR(?C)} \land \tdl{hasState(?C,$\bot$)} \land \tdl{hasTime(?C, ?t$_C$)}                              ~\land\notag\\
    &(\tdl{?t$_{O}$} \leqslant \tdl{?t$_C$})                                                                     ~\land\notag\\[.3em]
    &\tdl{CLEANED(?\vs{H})} \land \tdl{hasState(?\vs{H},$\top$)} \land \tdl{hasTime(?\vs{H}, ?t$_{\vs{H}}$)}    \land\notag\\
    &\tdl{CLEANED(?\vs{Q})} \land \tdl{hasState(?\vs{Q},$\top$)} \land \tdl{hasTime(?\vs{Q}, ?t$_{\vs{Q}}$)}    \land\notag\\   
    &(\tdl{?t$_{O}$} \leqslant \tdl{?t$_{\vs{H}}$}) \land (\tdl{?t$_{\vs{H}}$} \leqslant \tdl{?t$_C$})            \land
     (\tdl{?t$_{O}$} \leqslant \tdl{?t$_{\vs{Q}}$}) \land (\tdl{?t$_{\vs{Q}}$} \leqslant \tdl{?t$_C$})                 \notag\\
    &~~~~\implies \tdl{(\vs{A_7},$\top$):hasState} \land \tdl{(\vs{A_7},t$_C$):hasTime}. 
\end{align}
Similarly to Section~\ref{apx:sec:a3}, we compute the convolution through a computation performed by $\mathcal{M}_7$ before applying the rule.
Such operations are used to generate statements $\vs{H}$ and $\vs{O}$ in $\mathscr{T}_7$ as instances of the class \dltext{CLEANED}, which represent that a location has been cleaned.

\subsection{Activity Model \texorpdfstring{$\alpha_8$}{\textalpha\texteightinferior}}
\label{apx:sec:a8}

The eighth activity is choosing an outfit, and it involves $\mathscr{T}_8$.
The implementation based on the statement's algebra of the related model is
\begin{align}
    \label{eq:apx:a8}
    \vs{A}_8 \vDash \big(D^\top_{12} + \delta_{11}\big) &\leqslant \big(M^\top_{21} \lor M^\top_{22} \lor M^\top_{23}\big) \notag\\
                                &\leqslant \big(M^\top_3 \lor M^\top_4 \lor \ldots \lor M^\top_9\big).
\end{align}

\noindent
\vs{A}$_8$ can be expressed as 
\expl{a door was opened, then after $\delta_{11}$ time units, the person was in $M_{\{21:23\}}$ and then in $M_{\{3:9\}}$,}
where the door is associated with prior knowledge concerning the \dltext{CABINET2} (shown in Figures~\ref{fig:home} and \ref{apx:tab:onto}), and we consider 
\begin{align}
    \{\tdl{M$_{21}$},\ldots,\tdl{M$_{23}$}\}&\tdl{:CHOOSE}\sqsubseteq(\tdl{OUTFIT}\sqcap\tdl{MOTION}), \notag\\
          \{\tdl{M$_3$},\ldots,\tdl{M$_9$}\}&\tdl{:LEAVE}\sqsubseteq(\tdl{OUTFIT}\sqcap\tdl{MOTION}).
\end{align}
All those sensors concern the class \dltext{OUTFIT}, which is defined in $\mathscr{L}$ in accordance with \eqref{apx:eq:activitySensor} and Table~\ref{apx:tab:proc}.
Similarly to activity $\alpha_3$ (Section~\ref{apx:sec:a3}), we specify sub-areas in the corridor and in the living room, \ie where the person should \dltext{CHOOSE} the outfit and \dltext{LEAVE} it.
Differently from Section~\ref{apx:sec:a3}, the latter classes are not further specified as \emph{chosen} and \emph{left} since \vs{A_8} do not involve convolution, \ie no extra computation apart from OWL reasoning is required to compute \eqref{apx:sec:a8}.
The implementation of \eqref{eq:apx:a8} based on SWRL is the rule
\begin{align}
    &\tdl{DOOR(?O)} \land \tdl{hasState(?O,$\bot$)} \land \tdl{hasTime(?O, ?t$_O$)}                                                 ~\land\notag\\
    &\tdl{CHOOSE(?U)} \land \tdl{hasState(?U,$\top$)} \land \tdl{hasTime(?O, ?t$_U$)}                                               ~\land\notag\\
    &\tdl{LEAVE(?V)} \land \tdl{hasState(?V,$\top$)} \land \tdl{hasTime(?O, ?t$_V$)}                                                ~\land\notag\\
    &(\tdl{?t$_O$} \leqslant \tdl{?t$_U$}) \land (\tdl{?t$_O$} \leqslant \tdl{?t$_V$}) \land (\tdl{?t$_U$} \leqslant \tdl{?t$_V$})        \notag\\
    &~~~~\implies \tdl{(\vs{A_8},$\top$):hasState} \land \tdl{(\vs{A_8},t$_V$):hasTime}.
\end{align}

\end{document}